\documentclass{article} %
\usepackage{iclr2027_conference,times}

\usepackage{amsmath,amsfonts,bm}

\def\eqref#1{equation~\ref{#1}}

\def\1{\bm{1}}

\DeclareMathAlphabet{\mathsfit}{\encodingdefault}{\sfdefault}{m}{sl}
\SetMathAlphabet{\mathsfit}{bold}{\encodingdefault}{\sfdefault}{bx}{n}

\usepackage{booktabs}
\usepackage{graphicx}
\usepackage[caption=false,font=small]{subfig}
\usepackage{multirow}
\usepackage{float}
\usepackage{hyperref}
\usepackage{url}
\usepackage{booktabs}
\usepackage{tabularx}
\usepackage{array}
\usepackage[table]{xcolor}

\newcolumntype{C}{>{\centering\arraybackslash}X}

\title{\begin{minipage}[c]{0.82\linewidth}\raggedright
UniW(A)M: A Unified Mixture-of-Experts Framework for World and World-Action Models
\end{minipage}\hfill\begin{minipage}[c]{0.14\linewidth}\centering
\includegraphics[width=\linewidth]{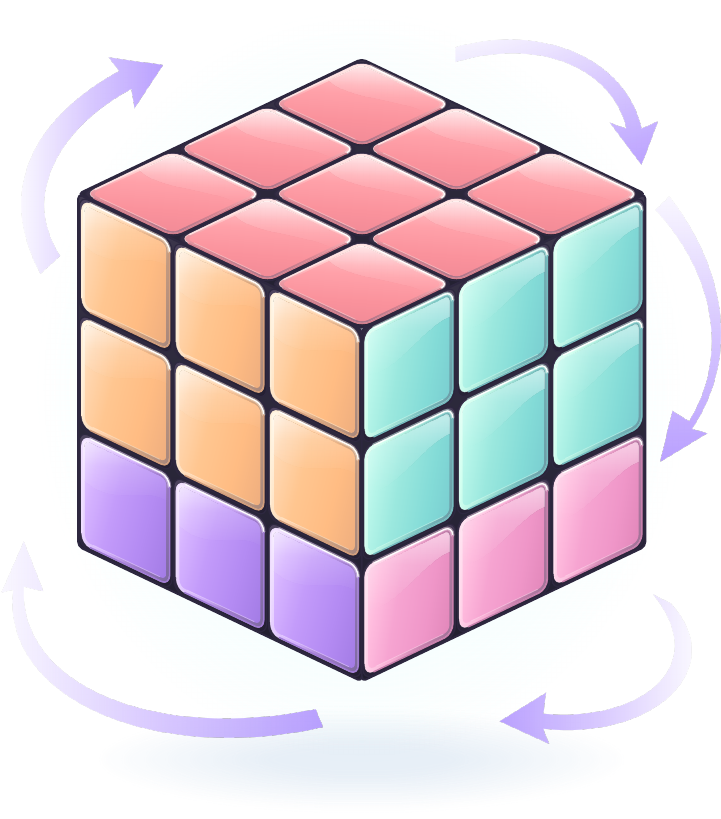}
\end{minipage}}

\author{%
\bfseries Jingbo Xu\textsuperscript{1,$*$}, Jianjie Fang\textsuperscript{1,$*$}, Zhuohang Li\textsuperscript{1}, Zihao Song\textsuperscript{1}, Zhaolu Wang\textsuperscript{2}, Xvyuan Liu\textsuperscript{3},\\
\bfseries Xin Zhang\textsuperscript{4}, Haisheng Su\textsuperscript{4}, Yinzhou Tang\textsuperscript{1}, Chen Gao\textsuperscript{1,2,$\dagger$}, Wei Wu\textsuperscript{4,$\dagger$}, Yong Li\textsuperscript{1,$\dagger$}\\[0.5em]
\normalfont\textsuperscript{1}Tsinghua University \quad \textsuperscript{2}Zhongguancun Academy\\
\textsuperscript{3}East China Normal University \quad \textsuperscript{4}Manifold AI\\[0.4em]
\textsuperscript{$*$}Joint first author. \quad \textsuperscript{$\dagger$}Corresponding authors.
}

\begin{document}

\maketitle

\begin{abstract}
World models (WMs) and world-action models (WAMs) focus on future world prediction and executable action generation, respectively. Although these models learn shared physical dynamics and interaction semantics from embodied trajectories, they are typically trained in isolation, which restricts knowledge reuse between prediction and control. The central challenge is therefore to develop a unified learning framework that shares these dynamics while preserving the distinct modeling capabilities required by each paradigm.
We propose \textbf{UniW(A)M}, a mixture-of-experts world-(action) model built upon diffusion Transformers that unifies action-conditioned world modeling and diverse WAM variants. Shared experts capture common dynamics, while atomic capability experts are reused across interfaces so that each interface keeps its own conditional pathway. At deployment, a plug-in adaptive interface router (PAIR) learns from a small subset of target-task chunk scenarios to select a suitable WAM interface for each execution chunk, balancing task success and execution efficiency while keeping the backbone frozen.
We evaluate UniW(A)M on LIBERO, RoboTwin 2.0, and three real-world manipulation tasks, where it achieves 88.3\% average real-world success versus 76.7\% for the strongest baseline. On WorldArena, it leads action-conditioned world modeling on all three tasks. AC-WM co-training contributes to action generation on every action interface and PAIR raises OOD success from 63.3\% to 76.7\%.

\end{abstract}

\section{Introduction}
\label{sec:intro}

\suppressfloats[t]
\begin{figure}[t]
\centering
\includegraphics[width=\linewidth]{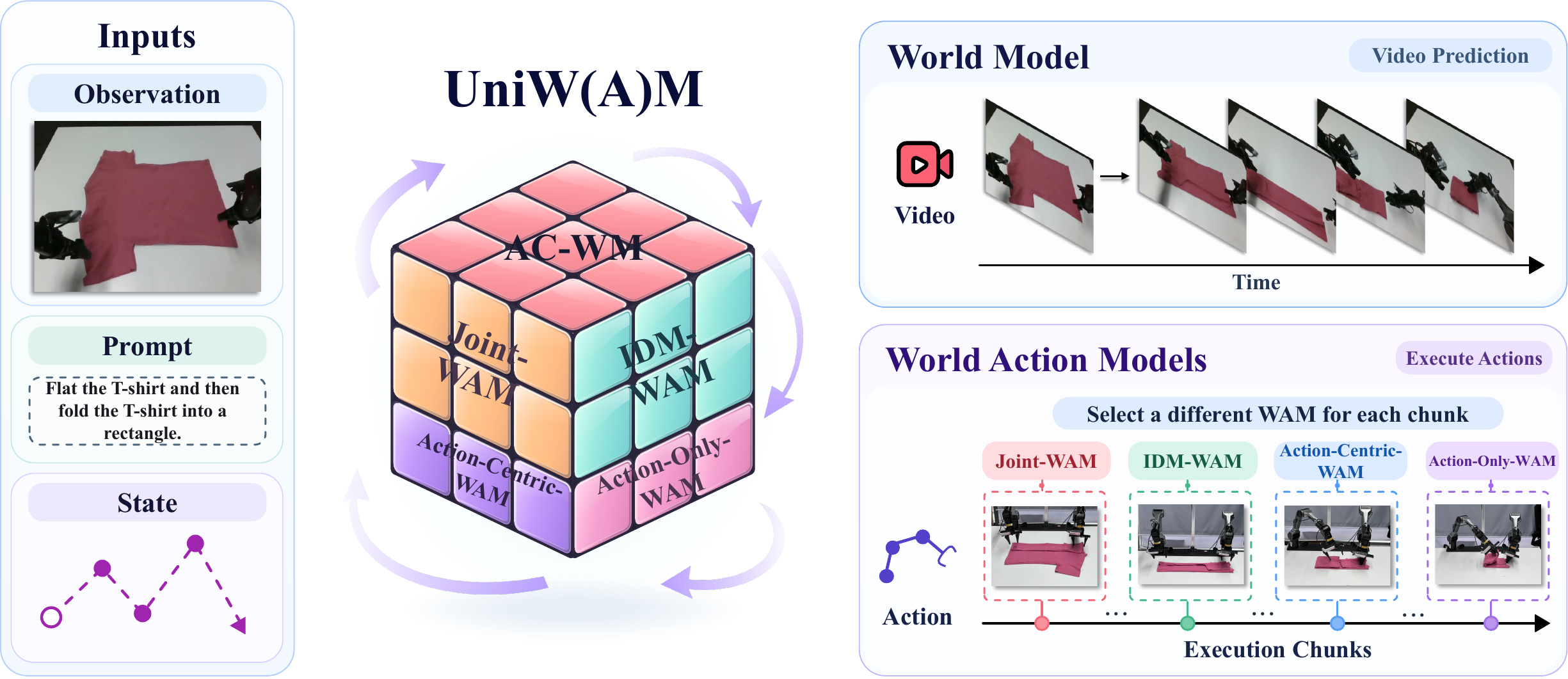}
\caption{Overview of UniW(A)M. The model unifies AC-WM and four WAM paradigms over shared embodied context. AC-WM predicts future video, while action execution adaptively switches among the four WAM paradigms across successive chunks, illustrated by a cloth-folding task.}
\label{fig:overview}
\end{figure}

World models (WMs) and world-action models (WAMs) offer complementary capabilities for embodied intelligence: predicting how environments evolve and generating actions to interact with them~\citep{260512090}. Action-conditioned world models simulate future observations given candidate actions~\citep{251010125}, whereas WAMs use future-world representations to support action generation~\citep{260620781}. WAM variants differ in how they use visual prediction, ranging from explicit or latent future modeling~\citep{260827033, 260809381} to action-only inference that omits future-video generation at deployment~\citep{260316666, 260317240}.

Existing unified models connect world prediction and action generation through shared generative modeling and modular architectures. Motus integrates understanding, video-generation, and action experts~\citep{251213030}, while UWM couples video and action diffusion to support multiple prediction modes~\citep{250402792}. Worldscape-MoE extends shared dynamics learning to heterogeneous action controls~\citep{260703964}. Together, these approaches demonstrate the value of shared modeling and modular specialization. For jointly trained video--action interfaces, the key design issue is how to reuse shared dynamics while organizing specialized computation around the predictive roles induced by different conditional dependencies.

Building on this direction, we jointly train action-conditioned world modeling and multiple WAM interfaces so that world modeling contributes to action generation and each interface keeps its own conditional pathway. These interfaces define different conditional relationships over the same embodied trajectories: future actions may be supplied as conditions or generated as outputs, and future video and action tokens interact under different visibility patterns. Video--action masks make these dependencies explicit within a common modeling framework.

We propose \textbf{UniW(A)M}, a mixture-of-experts framework that unifies action-conditioned world modeling and four WAM interfaces within a shared diffusion Transformer. Video--action visibility masks specify each interface's conditional dependencies. We organize the interfaces around four reusable predictive roles, which we term atomic capabilities: action-free video prediction, direct action generation, action-conditioned video prediction, and video-conditioned action generation. Shared experts model common dynamics, while atomic capability experts specialize in these roles and can be reused across interfaces. To align expert computation with these roles, Capability-Aligned Token Routing (\textbf{CATR}) assigns video and action target groups to eligible expert sets and learns mixture weights within each set. Joint training on the same trajectory data allows the interfaces to reuse these components while retaining their distinct conditional pathways.

These WAM interfaces differ in their use of future-video prediction and inference cost, raising a deployment question: which interface should be used at each execution chunk? The Plug-in Adaptive Interface Router (\textbf{PAIR}) learns this selection offline from a small subset of target-task chunk scenarios while keeping the UniW(A)M backbone frozen. At deployment, PAIR uses the current embodied context to select one of the four WAM interfaces for each chunk, allowing execution to switch between fast action-generation pathways that omit future-video prediction and interfaces that explicitly predict future video as interaction demands change.

On three real-world manipulation tasks, UniW(A)M achieves 88.3\% average success versus 76.7\% for the strongest baseline, and leads action-conditioned world modeling on all three WorldArena tasks. AC-WM co-training contributes to every action interface, including those that do not predict future video at deployment, and CATR keeps all four interfaces effective under joint training. Under distribution shifts, PAIR raises average success from 63.3\% to 76.7\% by selecting the interface for each chunk while the backbone stays frozen.

The main contributions of this work are summarized as follows:
\begin{itemize}
    \item We introduce \textbf{UniW(A)M}, a mixture-of-experts framework that unifies action-conditioned world modeling and four WAM interfaces through video–action visibility masks. Shared dynamics experts, atomic capability experts, and \textbf{CATR} enable joint training while each interface keeps its own conditional pathway.
    \item We develop \textbf{PAIR} for chunk-level WAM interface selection, adapting interface choice using a small subset of target-task chunk scenarios without updating the shared backbone, raising OOD success and improving execution efficiency.
    \item Experiments show that co-training contributes to action generation on every action interface and that the unified model leads action-conditioned world prediction, and that PAIR further raises OOD success by selecting the interface per chunk with the backbone frozen.
\end{itemize}

\section{Related Work}
\label{sec:related}

\paragraph{Action-conditioned world models.} Action-conditioned world models (AC-WMs) represent a branch of world models (WMs) dedicated to embodied robotic planning and evaluation~\citep{260916074}. Built primarily upon large-scale pre-trained video generative models, they simulate future temporal evolution conditioned on current observations and specified actions~\citep{251010125, 260604463, 240215391}. Unlike pure policy models, AC-WMs emphasize controllable action dependencies and focus primarily on future world prediction rather than joint optimization with action generation~\citep{260620781, 260824885, 260208971}. However, the general dynamics knowledge acquired through action-conditioned world modeling is rarely transferred to world action models, which limits overall data efficiency~\citep{260827033, 260711270, 260215922}.

\paragraph{World action models.} World action models (WAMs) integrate future world prediction and executable action generation within a unified learning objective, enabling action synthesis to exploit foresight into future states. Existing WAM variants differ in their generation targets and conditional structures~\citep{260620781, 260916074}. Specifically, joint interfaces predict video and actions concurrently~\citep{260215922}, visual-to-action interfaces synthesize visual futures prior to inferring actions~\citep{260527817}, and action-centric interfaces prioritize action output during inference while treating video primarily as a supervisory signal~\citep{260317240}. Although these variants rely on identical embodied trajectory data, their conditional dependencies, implemented via attention masks~\citep{260823927, 260810232, 260807468, 260500078}, vary considerably. Current methodologies typically train and deploy a single interface in isolation, which prevents the specialized action-generation capabilities developed across different interfaces from being shared and reused within a unified model~\citep{260827033, 260711270}.

\paragraph{Unified world-action modeling.} Mixture-of-Transformers (MoT) architectures, exemplified by Motus~\citep{251213030, 260427792}, integrate understanding, video generation, and action experts into a single framework, enabling flexible switching across multiple operational modes through coordinated scheduling. Another line of research employs a single Transformer backbone~\citep{260624101}, represented by UWM and UniWM, where UWM couples video and action diffusion via independent diffusion timesteps to facilitate flexible transitions among diverse modes~\citep{250402792}, while UniWM unifies planning and visual imagination within a multimodal autoregressive backbone~\citep{251008713}. A third direction adopts Mixture-of-Experts (MoE) architectures, such as Worldscape-MoE~\citep{260703964, 260620781}, which captures universal world dynamics via shared experts and processes heterogeneous control signals with condition-specific experts, thereby achieving modular disentanglement and scaling under multimodal action supervision. Nevertheless, the mutual reinforcement among distinct modes within such unified frameworks to improve overall model performance remains largely unexplored~\citep{260916074, 260326741}.

\section{Method}
\label{sec:method}

UniW(A)M jointly trains action-conditioned world modeling and four WAM interfaces on shared embodied trajectories, so that world modeling contributes to action generation while each interface keeps its own conditional pathway. Its design has three components. First, we formulate action-conditioned world modeling and four WAM variants as interfaces defined by video-action visibility masks, making their distinct conditional dependencies explicit within a unified framework. Second, these interfaces share a diffusion Transformer backbone and are jointly trained on the same trajectory data. Shared experts capture common dynamics, while CATR assigns each video and action token group to the experts of its predictive role. Third, PAIR selects one of the four WAM interfaces for each execution chunk. With the backbone frozen, it learns this selection from a small subset of target-task chunk scenarios, switching between fast action pathways and interfaces that predict future video under OOD conditions.

\subsection{Unified Modeling of Heterogeneous Video-Action Mask Relations}
\label{sec:method-interfaces}

An embodied trajectory comprises historical observations $o_{\le t}$, instructions $\ell$, future observations $o_{t+1:t+k}$, and future actions $a_{t+1:t+k}$. Interfaces differ in whether future actions are conditioning inputs or prediction targets, and in which future tokens can attend to one another during denoising. AC-WM predicts video conditioned on externally supplied actions; WAMs generate actions alongside video representations.

\paragraph{Atomic capabilities and paradigm compositions.}

\begin{table}[t]
\centering
\caption{Atomic capability composition and visibility characteristics of different paradigms.}
\label{tab:paradigms}
\begin{tabular}{lll}
\toprule
\textbf{Paradigm} & \textbf{Atomic Capabilities} & \textbf{Visibility / Role} \\
\midrule
AC-WM & WM & Action-conditioned world modeling \\
Joint-WAM & VGM + IDM & Mutual visibility \\
IDM-WAM & VGM + IDM & Sequential vision-to-action \\
Action-Centric-WAM & VLA + WM & Action-centric with video supervision \\
Action-Only-WAM & VLA + VGM & Action generation with auxiliary video \\
\bottomrule
\end{tabular}
\end{table}

We define four atomic capabilities: \textbf{VGM}, action-free video prediction; \textbf{VLA}, direct action generation; \textbf{WM}, action-conditioned video prediction; and \textbf{IDM}, video-conditioned action generation. VGM and WM model the future world, whereas VLA and IDM model future actions. These capabilities organize predictive roles within our chosen interfaces. Each expert receives the representation admitted by that interface's attention mask, and the role name refers to this masked prediction.

\suppressfloats[t]
\begin{figure}[t]
\centering
\includegraphics[width=0.9\linewidth]{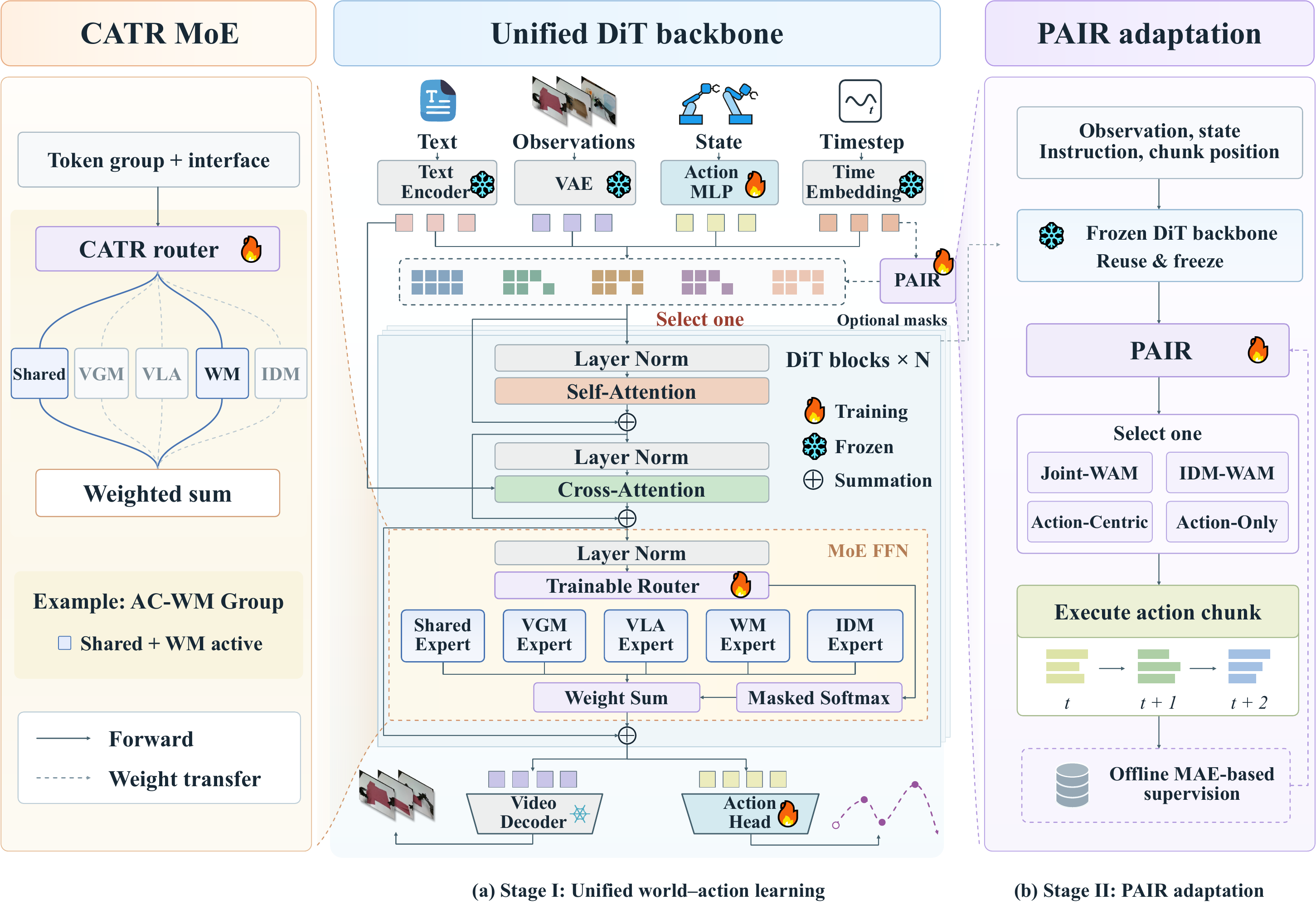}
\caption{Architecture of UniW(A)M. (a) Stage I: Unified learning. Five interfaces share a diffusion Transformer with CATR expert routing. (b) Stage II: PAIR adaptation. With the backbone frozen, PAIR learns interface selection from offline MAE-based supervision.}
\label{fig:moe}
\end{figure}

\paragraph{Conditional directions of the interfaces.}
Table~\ref{tab:paradigms} summarizes how the atomic capabilities are composed into five interfaces. Each interface specifies the role of future actions and the visibility between future video and action tokens.

\textbf{Joint-WAM} combines the video-prediction role of VGM with the video-conditioned action-prediction role of IDM, motivated by decomposing a joint video--action distribution into a video marginal and an action conditional, as in DreamZero~\citep{260215922}. The two modalities are denoised jointly with bidirectional attention, allowing their evolving representations to inform one another. \textbf{IDM-WAM} uses the same VGM and IDM capabilities with a sequential dependency. VGM first predicts a visual rollout, and IDM then infers actions conditioned on that rollout. Action tokens can access the predicted future video, while future video tokens do not attend to future actions. \textbf{Action-Centric-WAM} combines VLA and WM: VLA predicts actions without accessing future video, while WM predicts their visual consequences as auxiliary training supervision. \textbf{Action-Only-WAM} combines VLA and VGM with both directions of future cross-modal attention disabled, so the two branches predict their respective targets from the observed context. Both interfaces omit future-video prediction at deployment. \textbf{AC-WM} activates WM to predict future observations given candidate actions. These actions are clean external conditions rather than prediction targets, so the interface supports world simulation without generating executable actions.

The conditional probability formulations of all five interfaces are provided in Appendix~\ref{sec:appendix-interface-formulations}, and their training and inference visibility masks are illustrated in Appendix~\ref{sec:appendix-interface-masks}.

\subsection{Architecture of UniW(A)M}
\label{sec:method-moe}
\label{sec:method-train-selection}

As illustrated in Figure~\ref{fig:moe}, UniW(A)M uses interface-specific attention masks and replaces the diffusion Transformer's dense feed-forward sublayers with mixture-of-experts modules, initialized from the pretrained dense sublayers as described in Appendix~\ref{sec:appendix-expert-initialization}.

\paragraph{Heterogeneous Expert Activation Architecture.}
Shared experts participate in every interface, while atomic experts specialize in interface-specific predictive roles. The same expert ensemble can support different interfaces through distinct attention masks. Training and inference expert assignments are detailed in Appendix~\ref{sec:appendix-expert-activation}.

\paragraph{Capability-Aligned Token Routing.}
We introduce Capability-Aligned Token Routing (CATR) to align expert computation with the predictive roles of video and action token groups. Each target group is assigned a prescribed set of eligible experts based on its modality and interface-specific predictive role. A learned router produces group-level mixture weights within this set; experts outside the set do not contribute to the group's output. Tokens within a group share these weights but retain their individual expert outputs. Video target groups use VGM or WM specialists, whereas action target groups use VLA or IDM specialists, alongside shared experts. The eligible sets and treatment of clean conditions are detailed in Appendix~\ref{sec:appendix-expert-activation}. Guided by CATR, UniW(A)M is jointly optimized through the multi-task training objective detailed in Appendix~\ref{sec:appendix-training-objectives}.

\paragraph{Inference-Time Computation and Latency.}
IDM-WAM generates video and actions sequentially, whereas Joint-WAM denoises them jointly. Action-Centric-WAM and Action-Only-WAM omit future-video prediction and use shared and VLA experts for action denoising. Their conditioning paths differ: WM is inactive in Action-Centric-WAM, while Action-Only-WAM additionally uses VGM for the initial conditioning frame, allowing different contextual representations and action predictions. Detailed expert assignments and cache handling are provided in Appendix~\ref{sec:appendix-expert-activation}.

\subsection{Out-of-Distribution Deployment with Plug-in Adaptive Routing}
\label{sec:method-adaptive-router}

The UniW(A)M backbone is frozen after pretraining. To adapt interface selection to a target OOD task without changing the backbone, we train a lightweight Plug-in Adaptive Interface Router (\textbf{PAIR}) on a small subset of target-task chunk scenarios. PAIR uses a trainable classification head to select among the four WAM interfaces at each action chunk. Let $\mathcal{I}_{\mathrm{WAM}}$ denote these four action-generating interfaces in Table~\ref{tab:paradigms}: Joint-WAM, IDM-WAM, Action-Centric-WAM, and Action-Only-WAM.

During offline router training, each frozen WAM interface predicts actions for the same chunk scenario. We compare each prediction with the ground-truth actions using mean absolute error (MAE), then min--max normalize the four errors to obtain $\Delta_t^{(i)}$ for interface $i\in\mathcal{I}_{\mathrm{WAM}}$. The target probability decreases with this normalized error:
\begin{equation}
\tilde{y}_t^{(i)}
=
\frac{
\exp\left(-\Delta_t^{(i)}/\tau\right)
}{
\sum\limits_{j\in\mathcal{I}_{\mathrm{WAM}}}
\exp\left(-\Delta_t^{(j)}/\tau\right)
},
\qquad
i \in \mathcal{I}_{\mathrm{WAM}},
\label{eq:router-soft-target}
\end{equation}
Here $\tau>0$ controls the target sharpness. Among interfaces whose MAE is within 1\% of the minimum, an efficiency preference selects a fast interface if available, choosing the faster one if both qualify. The selected interface receives a one-hot target; otherwise, the MAE-based soft target is retained. We denote the resulting target by $y_t^{(i)}$; its construction is detailed in Appendix~\ref{sec:appendix-pair-targets}.

PAIR maps frozen contextual features $z_t$ to interface probabilities $q_{\phi}(i\mid z_t)$ with classification-head parameters $\phi$. It is trained by cross-entropy against the MAE-aware target:
\begin{equation}
\mathcal{L}_{\mathrm{Router}}
=
-\sum_{i\in\mathcal{I}_{\mathrm{WAM}}}
y_t^{(i)}
\log q_{\phi}(i\mid z_t),
\label{eq:router-loss}
\end{equation}
At deployment, PAIR chooses the interface with the highest predicted probability for each action chunk, executes that interface's actions, and updates the observation history before the next selection. Router features, output probabilities, and the selection rule are specified in Appendix~\ref{sec:appendix-pair-inference}.

\paragraph{Router inference cost.}
PAIR is invoked at chunk transitions. Its ensemble prediction takes approximately 2~ms; including shared-expert feature computation, routing averages approximately 63~ms in the benchmark described in Appendix~\ref{sec:appendix-pair-inference}, excluding observation and language encoding. The appendix specifies the measurement scope and cache handling.

\section{Experiments}
\label{sec:experiments}

\subsection{Experimental Setup}
\label{sec:exp-setup}
\begin{figure}[htbp]
\centering
\includegraphics[width=0.8\linewidth]{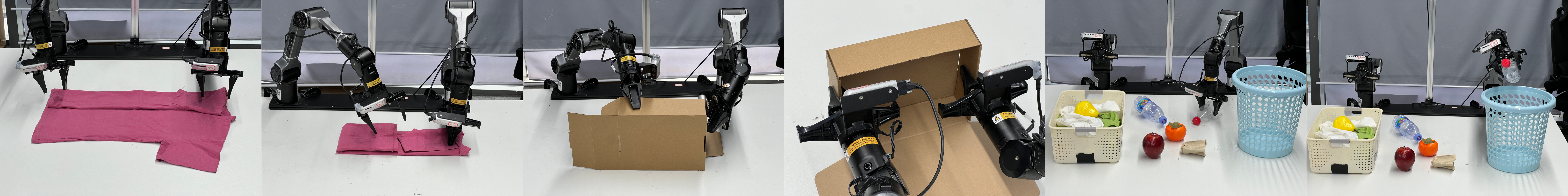}
\caption{Real-world tasks: cloth folding, box folding, and tabletop cleaning.}
\label{fig:real-world-tasks}
\end{figure}

To ensure fair and reproducible comparisons, we evaluate models under the same task settings and protocols, covering future-world modeling and action generation. During unified training, we sample the five paradigms uniformly and optimize one paradigm per step. We evaluate action generation on LIBERO~\citep{230603310}, RoboTwin 2.0~\citep{250618088}, and three real-world tasks: cloth folding, box folding, and tabletop cleaning (Figure~\ref{fig:real-world-tasks}). AC-WM's action-conditioned video prediction is evaluated with WorldArena~\citep{260208971}. Further details on the benchmarks and evaluation protocols are provided in Appendix~\ref{sec:appendix-benchmarks}.

Per-task success counts and 95\% Wilson intervals for the real-world evaluations are reported in Appendix~\ref{sec:appendix-uncertainty}.

\paragraph{OOD settings.} We retain the same task goals and define OOD relative to backbone training: unseen garment colors and sizes for cloth folding, different box types for box folding, and different cluttered backgrounds for tabletop cleaning.

\subsection{Overall Evaluation}
\label{sec:exp-main}

The main analysis focuses on the three real-world manipulation tasks, which also form the basis of our ablation studies. We additionally evaluate action generation on LIBERO and RoboTwin 2.0. UniW(A)M achieves 97.5\% average success on LIBERO, including 99.2\% on Spatial; detailed results are provided in Appendix~\ref{sec:appendix-simulation-results}.

\begin{figure}[t]
\centering
\subfloat[Clean-to-Clean (C2C)\label{fig:robotwin-c2c}]{\includegraphics[width=0.49\linewidth]{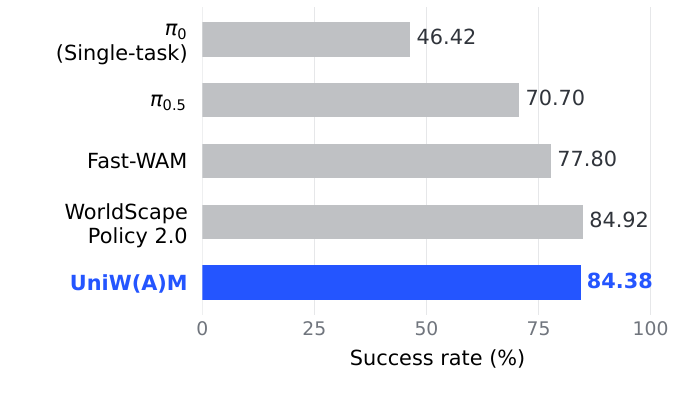}}
\hfill
\subfloat[Clean-to-Randomized (C2R)\label{fig:robotwin-c2r}]{\includegraphics[width=0.49\linewidth]{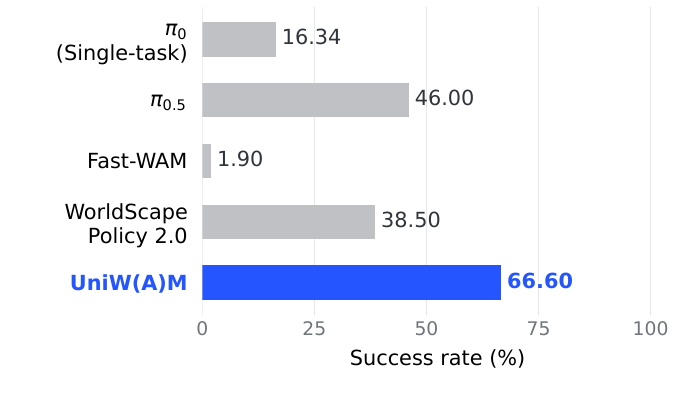}}
\caption{RoboTwin 2.0 success rates with clean-only training. \textbf{(a)} C2C evaluates on clean scenes; \textbf{(b)} C2R evaluates on randomized scenes. Scores are averaged over 50 tasks separately for each setting. Gray bars show official leaderboard baselines~\citep{robotwin-leaderboard-2026}: $\pi_0$ uses single-task fine-tuning, while the others use co-training. Blue bars show UniW(A)M.}
\label{fig:robotwin}
\end{figure}

\paragraph{RoboTwin generalization.}
With clean-only training, UniW(A)M achieves 84.38\% success under Clean-to-Clean (C2C) evaluation and 66.60\% under Clean-to-Randomized (C2R) evaluation (Figure~\ref{fig:robotwin}). It ranks second in C2C among the compared methods and leads in C2R, exceeding the strongest listed C2R baseline, $\pi_{0.5}$, by 20.60 percentage points. These results demonstrate effective generalization to randomized scenes without randomized training demonstrations. Protocol details and numerical comparisons are provided in Appendices~\ref{sec:appendix-benchmarks} and~\ref{sec:appendix-simulation-results}.

\paragraph{Results and analysis.}

On real-world evaluation, Table~\ref{tab:real} reports real-world success rates over 20 trials per task. All four UniW(A)M interfaces exceed the strongest baseline in average success, with IDM-WAM achieving 88.3\% versus 76.7\% for WorldScape Policy 2.0, a gain of approximately 11.7 percentage points. IDM-WAM reaches 90\%, 85\%, and 90\% on cloth folding, box folding, and tabletop cleaning, respectively. Qualitative visualizations of the three tasks across all four interfaces are provided in Appendix~\ref{sec:appendix-real-world-sequences}.

On WorldArena, UniW(A)M achieves the highest aggregate EWMScore-P among the compared methods on all three tasks (Table~\ref{tab:worldarena-quality}), so the unified model leads action-conditioned world modeling together with the action-generation results above. Complete component scores and qualitative comparisons with Ctrl-World and Cosmos-Predict2.5 are provided in Appendices~\ref{sec:appendix-worldarena-metrics} and~\ref{sec:appendix-world-prediction-qualitative}, respectively.

\begin{table}[t]
\centering
\footnotesize %
\setlength{\tabcolsep}{5.0pt} %
\renewcommand{\arraystretch}{1.05} %
\caption{Success rates (\%) on real-world robotic tasks under the standard evaluation configuration. The best results within each task are shown in \textbf{bold}.}
\label{tab:real}
\begin{tabular}{lcccc}
\toprule
\multirow{2}{*}{\textbf{Method}} & \textbf{Folding Clothes} & \textbf{Folding Box} & \textbf{Tabletop Clean.} & \multirow{2}{*}{\textbf{Average}} \\
& \scriptsize\textit{(Deformable)} & \scriptsize\textit{(Contact-Rich)} & \scriptsize\textit{(Long-Horizon)} & \\
\midrule
\rowcolor{gray!15}
\multicolumn{5}{c}{\textit{Baselines}} \\
$\pi_{0.5}$~\citep{250416054}             & 60 & 65 & 70 & 65.0 \\
DreamZero~\citep{260215922}               & 45 & 55 & 60 & 53.3 \\
WorldScape Policy 2.0~\citep{260718840}   & 75 & 75 & 80 & 76.7 \\
\midrule
\rowcolor{gray!15}
\multicolumn{5}{c}{\textit{UniW(A)M}} \\
Joint-WAM               & \textbf{90} & 80 & \textbf{90} & 86.7 \\
IDM-WAM                 & \textbf{90} & \textbf{85} & \textbf{90} & \textbf{88.3} \\
Action-Centric-WAM      & 85 & 75 & 85 & 81.7 \\
Action-Only-WAM         & 85 & 70 & 85 & 80.0 \\
\bottomrule
\end{tabular}
\end{table}

\begin{table*}[t]
\centering
\caption{Evaluation of action-driven and interaction metrics of WorldArena on three real-world tasks. All metrics are reported on a $0$--$100$ scale. The best and second-best results within each task are shown in \textbf{bold} and \underline{underlined}, respectively.}
\label{tab:worldarena-quality}
\footnotesize
\setlength{\tabcolsep}{7.5pt} %
\renewcommand{\arraystretch}{1.15}
\resizebox{0.95\textwidth}{!}{%
\begin{tabular}{lcccccc}
\toprule
\textbf{Model} 
& \begin{tabular}[c]{@{}c@{}}\textbf{JEPA}\\\textbf{Similarity}$\uparrow$\end{tabular} 
& \begin{tabular}[c]{@{}c@{}}\textbf{Motion}\\\textbf{Smoothness}$\uparrow$\end{tabular} 
& \begin{tabular}[c]{@{}c@{}}\textbf{Interaction}\\\textbf{Quality}$\uparrow$\end{tabular} 
& \begin{tabular}[c]{@{}c@{}}\textbf{Trajectory}\\\textbf{Accuracy}$\uparrow$\end{tabular} 
& \begin{tabular}[c]{@{}c@{}}\textbf{Instruction}\\\textbf{Following}$\uparrow$\end{tabular} 
& \begin{tabular}[c]{@{}c@{}}\textbf{EWM}\\\textbf{Score-P}$\uparrow$\end{tabular} \\
\midrule
\rowcolor{gray!15}
\multicolumn{7}{c}{\textit{Task 1: Folding Clothes (Deformable)}} \\
Cosmos-Predict 2.5~\citep{251100062} & \underline{21.86} & 60.70 & \underline{64.00} & 7.20 & \underline{56.50} & 48.08 \\
CtrlWorld~\citep{251010125}          & 17.57 & \underline{78.21} & 63.50 & \underline{18.56} & 51.50 & \underline{53.09} \\
\textbf{UniW(A)M (Ours)} & \textbf{44.39} & \textbf{87.77} & \textbf{80.00} & \textbf{22.87} & \textbf{93.50} & \textbf{61.51} \\
\midrule
\rowcolor{gray!15}
\multicolumn{7}{c}{\textit{Task 2: Folding Box (Contact-Rich)}} \\
Cosmos-Predict 2.5~\citep{251100062} & 40.27 & 52.47 & \textbf{76.50} & 10.91 & \textbf{95.00} & 51.07 \\
CtrlWorld~\citep{251010125}          & \underline{63.81} & \underline{66.56} & 64.50 & \underline{14.73} & 78.00 & \underline{55.48} \\
\textbf{UniW(A)M (Ours)} & \textbf{65.41} & \textbf{68.69} & \underline{70.50} & \textbf{17.28} & \underline{84.00} & \textbf{55.89} \\
\midrule
\rowcolor{gray!15}
\multicolumn{7}{c}{\textit{Task 3: Tabletop Clean. (Long-Horizon)}} \\
Cosmos-Predict 2.5~\citep{251100062} & 4.63 & 48.49 & 54.00 & 2.88 & 41.00 & 31.75 \\
CtrlWorld~\citep{251010125}          & \underline{33.76} & \underline{57.93} & \underline{60.00} & \underline{5.94} & \underline{59.00} & \underline{50.23} \\
\textbf{UniW(A)M (Ours)} & \textbf{50.30} & \textbf{63.10} & \textbf{66.50} & \textbf{7.26} & \textbf{68.50} & \textbf{52.53} \\
\bottomrule
\end{tabular}%
}
\end{table*}

\subsection{Analysis of Out-of-Distribution Scenarios}
\label{sec:exp-selector}

PAIR raises average OOD success from 63.3\% with the strongest standalone interface to 76.7\% (Table~\ref{tab:ood-real}), using routing-only adaptation with the UniW(A)M backbone frozen and 2,000 target-task chunk scenarios per task (Appendix~\ref{sec:appendix-pair-targets}).

Across Reach (approach), Contact (contact-based manipulation), and Core (fine manipulation), PAIR selects Action-Centric-WAM and Action-Only-WAM during approach, and Joint-WAM and IDM-WAM more often during contact and fine manipulation. Detailed selection distributions and critical-operation case studies are provided in Appendices~\ref{sec:appendix-pair-selection} and~\ref{sec:appendix-pair-case-study}, respectively.

\begin{table}[t]
\centering
\caption{OOD success rates (\%) for UniW(A)M interface adaptation on real-world tasks. The best results within each task are shown in \textbf{bold}.}
\label{tab:ood-real}
\footnotesize
\setlength{\tabcolsep}{5.2pt} %
\renewcommand{\arraystretch}{1.08}
\begin{tabular}{lcccc}
\toprule
\multirow{2}{*}{\textbf{Interface}} & \textbf{Folding Clothes} & \textbf{Folding Box} & \textbf{Tabletop Clean.} & \multirow{2}{*}{\textbf{Average}} \\
& \scriptsize\textit{(Deformable)} & \scriptsize\textit{(Contact-Rich)} & \scriptsize\textit{(Long-Horizon)} & \\
\midrule
\rowcolor{gray!15}
\multicolumn{5}{c}{\textit{Standalone Interfaces}} \\
Joint-WAM           & 65 & 55 & 70 & 63.3 \\
IDM-WAM             & 65 & 50 & 75 & 63.3 \\
Action-Centric-WAM  & 50 & 40 & 55 & 48.3 \\
Action-Only-WAM     & 50 & 45 & 55 & 50.0 \\
\midrule
\rowcolor{gray!10}
\textbf{PAIR} & \textbf{75} & \textbf{70} & \textbf{85} & \textbf{76.7} \\
\bottomrule
\end{tabular}
\end{table}

\subsection{Pareto Analysis of Success Rate and Inference Latency}
\label{sec:exp-pareto}

\begin{figure}[t]
\centering
\begin{minipage}[t]{0.49\linewidth}
\centering
\includegraphics[width=0.9\linewidth]{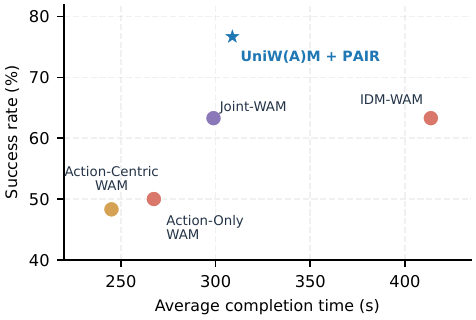}\\[2pt]
\textbf{(a) Task-level Pareto efficiency}
\end{minipage}
\hfill
\begin{minipage}[t]{0.49\linewidth}
\centering
\includegraphics[width=0.9\linewidth]{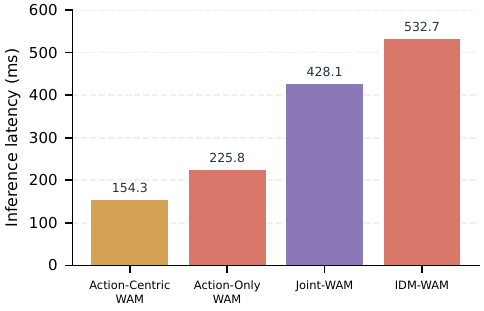}\\[2pt]
\textbf{(b) Step-normalized chunk latency}
\end{minipage}
\caption{Inference efficiency and task-level Pareto analysis of UniW(A)M. \textbf{(a)} Task success rate versus average completion time over successful real-robot trials, where UniW(A)M + PAIR occupies the upper-left Pareto frontier. \textbf{(b)} Full chunk inference time divided by the configured step count of 16 across the four standalone WAM interfaces; IDM-WAM uses 16 steps per stage (Appendix~\ref{sec:appendix-efficiency-metrics}). Chunk-boundary routing is excluded from these bars: ensemble prediction takes approximately 2~ms, while routing including shared-expert feature computation averages approximately 63~ms under the measurement scope in Appendix~\ref{sec:appendix-pair-inference}.}
\label{fig:latency-tradeoff}
\end{figure}

We measure the trade-off between task execution capability and inference cost for each interface. Figure~\ref{fig:latency-tradeoff}(b) reports full action-chunk inference time divided by the configured step count of 16: Action-Centric-WAM and Action-Only-WAM require 154.3~ms and 225.8~ms, respectively, whereas Joint-WAM and IDM-WAM require 428.1~ms and 532.7~ms. Appendix~\ref{sec:appendix-efficiency-metrics} specifies the timing metrics.
As shown in Figure~\ref{fig:latency-tradeoff}(a), UniW(A)M + PAIR achieves a 76.7\% success rate with an average completion time of 308.9~s over successful trials, placing it on the upper-left Pareto frontier of the four static interfaces. By routing to the fast action pathways during standard transfer motions and dynamically invoking foresight-capable joint pathways during contact-rich phases, the method combines the execution efficiency of action-centric interfaces with the foresight of video-conditioned interfaces.

\subsection{Ablation Studies}
\label{sec:exp-ablation}

Per-task results for the architecture, routing, and training-paradigm ablations are provided in Appendix~\ref{sec:appendix-ablation}.

\paragraph{Contribution of AC-WM co-training.} Removing AC-WM from mixed training reduces success across all four WAM interfaces on all three tasks. Average success decreases by 15.0 percentage points for both Joint-WAM and IDM-WAM, and by 31.7 and 43.3 points for Action-Centric-WAM and Action-Only-WAM, respectively. The larger drops on Action-Centric-WAM and Action-Only-WAM show that AC-WM co-training contributes to action generation even on interfaces that do not predict future video at deployment. Detailed results are provided in Appendix~\ref{sec:appendix-architecture-ablation}.

\paragraph{Architecture and training configuration.} We compare a single dense DiT initialized from Wan2.1-I2V-14B~\citep{250320314} with joint UniW(A)M training under the same training steps, learning-rate settings, and data volume; the detailed results are reported in Appendix~\ref{sec:appendix-architecture-ablation}. All four interfaces of the jointly trained UniW(A)M substantially outperform the baseline, whose best success rate is only 16.7\%, whereas UniW(A)M achieves 88.3\%.

\paragraph{CATR and sample-level routing.} The two MoE configurations have identical expert counts, network structures, and parameter counts. They differ in routing organization: sample-level routing applies the active expert ensemble to the entire sample, whereas CATR assigns eligible expert sets and learned mixture weights to modality- and role-specific token groups. This restricts specialist FFN computation to the relevant groups while retaining shared attention and shared experts. As shown in Appendix~\ref{sec:appendix-architecture-ablation}, sample-level routing produces uneven performance: Joint-WAM and IDM-WAM achieve 51.7\% and 60.0\% average success, respectively, whereas both Action-Centric-WAM and Action-Only-WAM achieve only 10.0\%. With expert count and parameter count held fixed, CATR improves all four interfaces and raises Action-Centric-WAM and Action-Only-WAM from 10.0\% to 81.7\% and 80.0\%, so joint training delivers all four interfaces.

\section{Conclusion}
\label{sec:conclusion}

UniW(A)M unifies action-conditioned world modeling and four WAM interfaces through video--action masks and capability-aligned expert routing. Experiments demonstrate action generation and future-world prediction within a shared backbone. Ablations show that AC-WM co-training improves all four WAM interfaces across three real-world tasks, while CATR improves action generation relative to sample-level routing. PAIR further improves OOD execution through chunk-level interface selection with the backbone frozen. Together, these results support joint learning and adaptive deployment of complementary world and action modeling capabilities.

\subsection*{AI use statement}

We used generative artificial intelligence tools to assist with drafting and editing the manuscript text, organizing and debugging LaTeX source files, and preparing schematic figures and plotting scripts. These tools were not used to generate experimental data, fabricate results, or substitute for author decisions regarding research methodology. All AI-assisted text, equations, code, and figures were reviewed, verified, and revised by the authors. We take full responsibility for the final content of this work, including all claims, analyses, and artifacts developed with generative artificial intelligence assistance.

\subsection*{Reproducibility statement}

To support reproducibility, the implementation complements the method descriptions and training details provided in the paper and appendix.

\bibliography{iclr2027_conference}

@misc{robotwin-leaderboard-2026,
  title = {{RoboTwin 2.0 Leaderboard}},
  author = {{RoboTwin Team}},
  year = {2026},
  url = {https://robotwin-platform.github.io/leaderboard},
  note = {Co-train results; updated September 24, 2026; accessed September 26, 2026}
}

@article{241024164,
  title = {{$\pi_0$}: A Vision-Language-Action Flow Model for General Robot Control},
  author = {Kevin Black and Noah Brown and Danny Driess and Adnan Esmail and Michael Equi and Chelsea Finn and others},
  journal = {arXiv preprint arXiv:2410.24164},
  year = {2024},
  eprint = {2410.24164},
  archivePrefix = {arXiv},
  doi = {10.48550/arXiv.2410.24164},
  url = {https://arxiv.org/abs/2410.24164}
}

@article{260121998,
  title = {Causal World Modeling for Robot Control},
  author = {Lin Li and Qihang Zhang and Yiming Luo and Shuai Yang and Ruilin Wang and Fei Han and others},
  journal = {arXiv preprint arXiv:2601.21998},
  year = {2026},
  eprint = {2601.21998},
  archivePrefix = {arXiv},
  doi = {10.48550/arXiv.2601.21998},
  url = {https://arxiv.org/abs/2601.21998}
}

@article{260316666,
  title = {{Fast-WAM}: Do World Action Models Need Test-time Future Imagination?},
  author = {Tianyuan Yuan and Zibin Dong and Yicheng Liu and Hang Zhao},
  journal = {arXiv preprint arXiv:2603.16666},
  year = {2026},
  eprint = {2603.16666},
  archivePrefix = {arXiv},
  doi = {10.48550/arXiv.2603.16666},
  url = {https://arxiv.org/abs/2603.16666}
}

@article{251100062,
  title = {World Simulation with Video Foundation Models for Physical {AI}},
  author = {{NVIDIA} and Arslan Ali and Junjie Bai and Maciej Bala and Yogesh Balaji and Aaron Blakeman and others},
  journal = {arXiv preprint arXiv:2511.00062},
  year = {2025},
  eprint = {2511.00062},
  archivePrefix = {arXiv},
  doi = {10.48550/arXiv.2511.00062},
  url = {https://arxiv.org/abs/2511.00062}
}

@article{260916074,
  title         = {World-Action Models for Robot Learning and Control: A Survey},
  author        = {Zuxing Lu and Hongjia Zhai and Guanzhi Wang and Huajian Zeng and Jiaqi Yang and Jingyu Liu and Lei Cheng and Yuantai Zhang and Yuheng Qiu and Zezhou Cheng and Ivan Laptev and Danfei Xu and Benjamin Riviere and Giuseppe Loianno and Eric Xing and Xingxing Zuo},
  journal       = {arXiv preprint arXiv:2609.16074},
  year          = {2026},
  eprint        = {2609.16074},
  archivePrefix = {arXiv},
  primaryClass  = {cs.RO}
}

@article{260620781,
  title         = {World Action Models: A Survey},
  author        = {Qiuhong Shen and Shihua Zhang and Yue Liao and Qi Li and Zhenxiong Tan and Shizun Wang and Shuicheng Yan and Xinchao Wang},
  journal       = {arXiv preprint arXiv:2606.20781},
  year          = {2026},
  eprint        = {2606.20781},
  archivePrefix = {arXiv},
  primaryClass  = {cs.RO}
}

@article{260527817,
  title         = {Turning Video Models into Generalist Robot Policies},
  author        = {Sizhe Lester Li and Evan Kim and Xingjian Bai and Tong Zhao and Tao Pang and Max Simchowitz and Vincent Sitzmann},
  journal       = {arXiv preprint arXiv:2605.27817},
  year          = {2026},
  eprint        = {2605.27817},
  archivePrefix = {arXiv},
  primaryClass  = {cs.RO}
}

@article{260317240,
  title         = {{GigaWorld-Policy}: An Efficient Action-Centered World--Action Model},
  author        = {Angen Ye and Boyuan Wang and Chaojun Ni and Guan Huang and Guosheng Zhao and Hao Li and Hengtao Li and Jie Li and Jindi Lv and Jingyu Liu and Min Cao and Peng Li and Qiuping Deng and Wenjun Mei and Xiaofeng Wang and Xinze Chen and Xinyu Zhou and Yang Wang and Yifan Chang and Yifan Li and Yukun Zhou and Yun Ye and Zhichao Liu and Zheng Zhu},
  journal       = {arXiv preprint arXiv:2603.17240},
  year          = {2026},
  eprint        = {2603.17240},
  archivePrefix = {arXiv},
  primaryClass  = {cs.CV}
}

@article{250320314,
  title = {Wan: Open and Advanced Large-Scale Video Generative Models},
  author = {{Team Wan} and Ang Wang and Baole Ai and Bin Wen and Chaojie Mao and others},
  journal = {arXiv preprint arXiv:2503.20314},
  year = {2025},
  eprint = {2503.20314},
  archivePrefix = {arXiv},
  doi = {10.48550/arXiv.2503.20314},
  url = {https://arxiv.org/abs/2503.20314}
}

@article{260215922,
  title         = {World Action Models are Zero-shot Policies},
  author        = {Seonghyeon Ye and Yunhao Ge and Kaiyuan Zheng and Shenyuan Gao and Sihyun Yu and George Kurian and Suneel Indupuru and You Liang Tan and Chuning Zhu and Jiannan Xiang and Ayaan Malik and Kyungmin Lee and William Liang and Nadun Ranawaka and Jiasheng Gu and Yinzhen Xu and Guanzhi Wang and Fengyuan Hu and Avnish Narayan and Johan Bjorck and Jing Wang and Gwanghyun Kim and Dantong Niu and Ruijie Zheng and Yuqi Xie and Jimmy Wu and Qi Wang and Ryan Julian and Danfei Xu and Yilun Du and Yevgen Chebotar and Scott Reed and Jan Kautz and Yuke Zhu and Linxi {``Jim''} Fan and Joel Jang},
  journal       = {arXiv preprint arXiv:2602.15922},
  year          = {2026},
  eprint        = {2602.15922},
  archivePrefix = {arXiv},
  primaryClass  = {cs.RO}
}

@article{260823927,
  title         = {{GlanceWAM}: Sparse Test-Time Imagination for World-Action Models},
  author        = {Linhan Wang and Zijian An and Mingyuan Zhang and Chen Dai and Yi Xu and Can Cui and Zichong Yang and Yinlin Chen and Lifeng Zhou and Chang-Tien Lu},
  journal       = {arXiv preprint arXiv:2608.23927},
  year          = {2026},
  eprint        = {2608.23927},
  archivePrefix = {arXiv},
  primaryClass  = {cs.CV}
}

@article{260810232,
  title         = {{FACT}: Failure-Aware Causal Training for World-Action Models},
  author        = {Quanquan Peng and Yutong Liang and Rui Yan and Nicklas Hansen and Xiaolong Wang},
  journal       = {arXiv preprint arXiv:2608.10232},
  year          = {2026},
  eprint        = {2608.10232},
  archivePrefix = {arXiv},
  primaryClass  = {cs.RO}
}

@article{260807468,
  title         = {{SimWAM}: A Simple World Action Model for End-to-End Autonomous Driving},
  author        = {Zongchuang Zhao and Xin Zhou and Tianyang Xu and Zhengyang Sun and Kaixuan Zhou and Yu Wu and Honglin Li and Dingkang Liang and Xiang Bai},
  journal       = {arXiv preprint arXiv:2608.07468},
  year          = {2026},
  eprint        = {2608.07468},
  archivePrefix = {arXiv},
  primaryClass  = {cs.CV}
}

@article{260500078,
  title         = {{Being-H0.7}: A Latent World-Action Model from Egocentric Videos},
  author        = {Hao Luo and Wanpeng Zhang and Yicheng Feng and Sipeng Zheng and Haiweng Xu and Chaoyi Xu and Ziheng Xi and Yuhui Fu and Zongqing Lu},
  journal       = {arXiv preprint arXiv:2605.00078},
  year          = {2026},
  eprint        = {2605.00078},
  archivePrefix = {arXiv},
  primaryClass  = {cs.RO}
}

@article{260827033,
  title         = {{Riemann-1.0}: An Embodied World Action Model for Physical {AI}},
  author        = {Haofeng Sun and Jiangbo Pei and Fei Kang and Zexiang Liu and Yaokun Li and Boyi Jiang and Hua Xue and Cindy Zhou and Wei Li and Yichen Wei and Mengyin An and Fanliang Zhao and Biao Jiang and Zile Wang and Yang Liu and Yangguang Li},
  journal       = {arXiv preprint arXiv:2608.27033},
  year          = {2026},
  eprint        = {2608.27033},
  archivePrefix = {arXiv},
  primaryClass  = {cs.RO}
}

@article{260711270,
  title         = {Towards Predictive, Aligned, and Scalable Robot Learning},
  author        = {Peijun Tang and Shangjin Xie and Baifu Huang and Binyan Sun and Haotian Yang and Kuncheng Luo and Weiqi Jin and Shilin Fang and Jianan Wang},
  journal       = {arXiv preprint arXiv:2607.11270},
  year          = {2026},
  eprint        = {2607.11270},
  archivePrefix = {arXiv},
  primaryClass  = {cs.RO}
}

@article{260427792,
  title         = {Motubrain: An Advanced World Action Model for Robot Control},
  author        = {{Motubrain Team} and Chendong Xiang and Fan Bao and Haitian Liu and Hengkai Tan and Hongzhe Bi and James Li and Jiabao Liu and Jingrui Pang and Kiro Jing and Louis Liu and Mengchen Cai and Rongxu Cui and Ruowen Zhao and Runqing Wang and Shuhe Huang and Yao Feng and Yinze Rong and Zeyuan Wang and Jun Zhu},
  journal       = {arXiv preprint arXiv:2604.27792},
  year          = {2026},
  eprint        = {2604.27792},
  archivePrefix = {arXiv},
  primaryClass  = {cs.RO}
}

@article{251213030,
  title         = {Motus: A Unified Latent Action World Model},
  author        = {Hongzhe Bi and Hengkai Tan and Shenghao Xie and Zeyuan Wang and Shuhe Huang and Haitian Liu and Ruowen Zhao and Yao Feng and Chendong Xiang and Yinze Rong and Hongyan Zhao and Hanyu Liu and Zhizhong Su and Lei Ma and Hang Su and Jun Zhu},
  journal       = {arXiv preprint arXiv:2512.13030},
  year          = {2025},
  eprint        = {2512.13030},
  archivePrefix = {arXiv},
  primaryClass  = {cs.CV}
}

@article{260624101,
  title         = {NavWM: A Unified Navigation World Model for Foresight-Driven Planning},
  author        = {Yanghong Mei and Longteng Guo and Ming-Ming Yu and Guiyu Zhao and Xingjian He and Jing Liu},
  journal       = {arXiv preprint arXiv:2606.24101},
  year          = {2026},
  eprint        = {2606.24101},
  archivePrefix = {arXiv},
  primaryClass  = {cs.RO}
}

@article{251008713,
  title         = {Towards Unified World Models for Visual Navigation via Memory-Augmented Planning and Foresight},
  author        = {Yifei Dong and Fengyi Wu and Guangyu Chen and Lingdong Kong and Qiyu Hu and Yuxuan Zhou and Xu Zhu and Jingdong Sun and Jun-Yan He and Qi Dai and Alexander G. Hauptmann and Zhi-Qi Cheng},
  journal       = {arXiv preprint arXiv:2510.08713},
  year          = {2025},
  eprint        = {2510.08713},
  archivePrefix = {arXiv},
  primaryClass  = {cs.AI}
}

@inproceedings{250402792,
  title         = {Unified World Models: Coupling Video and Action Diffusion for Pretraining on Large Robotic Datasets},
  author        = {Chuning Zhu and Raymond Yu and Siyuan Feng and Benjamin Burchfiel and Paarth Shah and Abhishek Gupta},
  booktitle     = {Robotics: Science and Systems (RSS)},
  year          = {2025},
  eprint        = {2504.02792},
  archivePrefix = {arXiv},
  primaryClass  = {cs.RO}
}

@article{260703964,
  title         = {Worldscape-MoE: A Unified Mixture-of-Experts World Model for Scalable Heterogeneous Action Control},
  author        = {Jianjie Fang and Yongyan Xu and Ziyou Wang and Chen Gao and Yuchao Huang and Zhaolu Wang and Rongze Tang and Mingyuan Jia and Baining Zhao and Weichen Zhang and Xin Zhang and Haisheng Su and Yu Shang and Wei Wu and Xinlei Chen and Yong Li},
  journal       = {arXiv preprint arXiv:2607.03964},
  year          = {2026},
  eprint        = {2607.03964},
  archivePrefix = {arXiv},
  primaryClass  = {cs.RO}
}

@article{260326741,
  title         = {Language-Conditioned World Modeling for Visual Navigation},
  author        = {Yifei Dong and Fengyi Wu and Yilong Dai and Lingdong Kong and Guangyu Chen and Xu Zhu and Qiyu Hu and Tianyu Wang and Johnalbert Garnica and Feng Liu and Siyu Huang and Qi Dai and Zhi-Qi Cheng},
  journal       = {arXiv preprint arXiv:2603.26741},
  year          = {2026},
  eprint        = {2603.26741},
  archivePrefix = {arXiv},
  primaryClass  = {cs.CV}
}

@article{260604463,
  title         = {{OSCAR}: Omni-Embodiment Action-Conditioned World Model for Robotics},
  author        = {Zhuoyuan Wu and Jun Gao},
  journal       = {arXiv preprint arXiv:2606.04463},
  year          = {2026},
  eprint        = {2606.04463},
  archivePrefix = {arXiv},
  primaryClass  = {cs.RO}
}

@article{251010125,
  title         = {Ctrl-World: A Controllable Generative World Model for Robot Manipulation},
  author        = {Yanjiang Guo and Lucy Xiaoyang Shi and Jianyu Chen and Chelsea Finn},
  journal       = {arXiv preprint arXiv:2510.10125},
  year          = {2025},
  eprint        = {2510.10125},
  archivePrefix = {arXiv},
  primaryClass  = {cs.RO}
}

@inproceedings{240215391,
  title         = {Genie: Generative Interactive Environments},
  author        = {Jake Bruce and Michael Dennis and Ashley Edwards and Jack Parker-Holder and Yuge Shi and Edward Hughes and Matthew Lai and Aditi Mavalankar and Richie Steigerwald and Chris Apps and Yusuf Aytar and Sarah Bechtle and Feryal Behbahani and Stephanie Chan and Nicolas Heess and Lucy Gonzalez and Simon Osindero and Sherjil Ozair and Scott Reed and Jingwei Zhang and Konrad Zolna and Jeff Clune and Nando de Freitas and Satinder Singh and Tim Rockt{\"a}schel},
  booktitle     = {International Conference on Machine Learning (ICML)},
  year          = {2024},
  eprint        = {2402.15391},
  archivePrefix = {arXiv},
  primaryClass  = {cs.LG}
}

@article{260824885,
  title         = {Do Robotic World Models Really Follow Actions? Diagnosing and Aligning Action-Conditioned Generation for Policy Learning},
  author        = {Sixiang Chen and Jiaming Liu and Jixian Wu and Yichen Guo and Tinghao Wang and Siyuan Qian and Hao Chen and Jiajun Cao and Jian Tang and Shanghang Zhang},
  journal       = {arXiv preprint arXiv:2608.24885},
  year          = {2026},
  eprint        = {2608.24885},
  archivePrefix = {arXiv},
  primaryClass  = {cs.RO}
}

@article{230603310,
  title         = {{LIBERO}: Benchmarking Knowledge Transfer for Lifelong Robot Learning},
  author        = {Bo Liu and Yifeng Zhu and Chongkai Gao and Yihao Feng and Qiang Liu and Yuke Zhu and Peter Stone},
  journal       = {arXiv preprint arXiv:2306.03310},
  year          = {2023},
  eprint        = {2306.03310},
  archivePrefix = {arXiv},
  primaryClass  = {cs.AI}
}

@article{250618088,
  title         = {{RoboTwin 2.0}: A Scalable Data Generator and Benchmark with Strong Domain Randomization for Robust Bimanual Robotic Manipulation},
  author        = {Tianxing Chen and Zanxin Chen and Baijun Chen and Zijian Cai and Yibin Liu and Zixuan Li and Qiwei Liang and Xianliang Lin and Yiheng Ge and Zhenyu Gu and Weiliang Deng and Yubin Guo and Tian Nian and Xuanbing Xie and Qiangyu Chen and Kailun Su and Tianling Xu and Guodong Liu and Mengkang Hu and Huan-ang Gao and Kaixuan Wang and Zhixuan Liang and Yusen Qin and Xiaokang Yang and Ping Luo and Yao Mu},
  journal       = {arXiv preprint arXiv:2506.18088},
  year          = {2025},
  eprint        = {2506.18088},
  archivePrefix = {arXiv},
  primaryClass  = {cs.RO}
}

@article{260208971,
  title         = {{WorldArena}: A Unified Benchmark for Evaluating Perception and Functional Utility of Embodied World Models},
  author        = {Yu Shang and Zhuohang Li and Yiding Ma and Weikang Su and Xin Jin and Ziyou Wang and Lei Jin and Xin Zhang and Yinzhou Tang and Haisheng Su and Chen Gao and Wei Wu and Xihui Liu and Dhruv Shah and Zhaoxiang Zhang and Zhibo Chen and Jun Zhu and Yonghong Tian and Tat-Seng Chua and Wenwu Zhu and Yong Li},
  journal       = {arXiv preprint arXiv:2602.08971},
  year          = {2026},
  eprint        = {2602.08971},
  archivePrefix = {arXiv},
  primaryClass  = {cs.CV}
}

@article{260512090,
  title         = {World Action Models: The Next Frontier in Embodied {AI}},
  author        = {Siyin Wang and Junhao Shi and Zhaoyang Fu and Xinzhe He and Feihong Liu and Chenchen Yang and Yikang Zhou and Zhaoye Fei and Jingjing Gong and Jinlan Fu and Mike Zheng Shou and Xuanjing Huang and Xipeng Qiu and Yu-Gang Jiang},
  journal       = {arXiv preprint arXiv:2605.12090},
  year          = {2026},
  eprint        = {2605.12090},
  archivePrefix = {arXiv},
  primaryClass  = {cs.RO}
}

@article{260809381,
  title         = {{JEPA-WAM}: Learning Vision-Language-Action Policies with Joint-Embedding World Modeling},
  author        = {Yihan Lin and Jiawei He and Shifeng Bao and Chen Zhao and Yang Li and Xiaobo Wang and Yan Wang and Cheng Chi and Jing Zhang},
  journal       = {arXiv preprint arXiv:2608.09381},
  year          = {2026},
  eprint        = {2608.09381},
  archivePrefix = {arXiv},
  primaryClass  = {cs.RO}
}

@article{260718840,
  title         = {{WorldScape Policy 2.0}: Empowering Steerable World Action Modeling with Reasoning-Augmented Memory and In-Context Learning},
  author        = {Haisheng Su and Zongdai Liu and Xin Jin and Haoxuan Dou and Chengming Hu and Baorun Li and Zhanwang Liu and Ruiyan Xu and Jianjie Fang and Xin Zhang and Zhenjie Yang and Xue Yang and Chen Gao and Junchi Yan and Yong Li and Wei Wu},
  journal       = {arXiv preprint arXiv:2607.18840},
  year          = {2026},
  eprint        = {2607.18840},
  archivePrefix = {arXiv},
  primaryClass  = {cs.RO}
}

@article{250416054,
  title         = {{$\pi_{0.5}$}: A Vision-Language-Action Model with Open-World Generalization},
  author        = {{Physical Intelligence} and Kevin Black and Noah Brown and James Darpinian and Karan Dhabalia and Danny Driess and Adnan Esmail and Michael Equi and Chelsea Finn and Niccolo Fusai and Manuel Y. Galliker and Dibya Ghosh and Lachy Groom and Karol Hausman and Brian Ichter and Szymon Jakubczak and Tim Jones and Liyiming Ke and Devin LeBlanc and Sergey Levine and Adrian Li-Bell and Mohith Mothukuri and Suraj Nair and Karl Pertsch and Allen Z. Ren and Lucy Xiaoyang Shi and Laura Smith and Jost Tobias Springenberg and Kyle Stachowicz and James Tanner and Quan Vuong and Homer Walke and Anna Walling and Haohuan Wang and Lili Yu and Ury Zhilinsky},
  journal       = {arXiv preprint arXiv:2504.16054},
  year          = {2025},
  eprint        = {2504.16054},
  archivePrefix = {arXiv},
  primaryClass  = {cs.LG}
}

@article{wilson1927probable,
  author = {Wilson, Edwin B.},
  title = {Probable Inference, the Law of Succession, and Statistical Inference},
  journal = {Journal of the American Statistical Association},
  volume = {22},
  number = {158},
  pages = {209--212},
  year = {1927},
  doi = {10.1080/01621459.1927.10502953}
}

@article{240609246,
  title={OpenVLA: An Open-Source Vision-Language-Action Model},
  author={Kim, Moo Jin and Pertsch, Karl and Karamcheti, Siddharth and Xiao, Ted and Balakrishna, Ashwin and Nair, Suraj and Rafailov, Rafael and Foster, Ethan and Lam, Grace and Sanketi, Pannag and Vuong, Quan and Kollar, Thomas and Burchfiel, Benjamin and Tedrake, Russ and Sadigh, Dorsa and Levine, Sergey and Liang, Percy and Finn, Chelsea},
  journal={arXiv preprint arXiv:2406.09246},
  year={2024}
}
\bibliographystyle{iclr2027_conference}

\appendix
\section{Video--Action Interfaces and Expert Routing}
\label{sec:appendix-interface-details}

\subsection{Conditional Formulations of the Five Interfaces}
\label{sec:appendix-interface-formulations}

We formalize the five interfaces in the same order as Section~\ref{sec:method-interfaces}. Here $o_t$ is the current observation, $o_{\le t}$ denotes the observation history, $\ell$ is the language instruction, and $o_{t+1:t+k}$ and $a_{t+1:t+k}$ are the future visual rollout and action sequence. Let $C_t^{(p)}$ denote the observed context available to interface $p$, including its visible observation history, language instruction, and robot-state inputs where used; AC-WM omits state tokens. We write $C_t$ for this interface-specific context below. Future video and action targets are kept explicit. The factors express modeled conditional relationships, while attention masks and denoising schedules specify their realization.

\textbf{Joint-WAM.} The joint distribution over the two future modalities can be written as a video marginal and an action conditional, corresponding to the VGM and IDM capabilities:
\begin{equation}
p(o_{t+1:t+k}, a_{t+1:t+k}\mid C_t)
=
p_{\mathbf{VGM}}(o_{t+1:t+k}\mid C_t)
p_{\mathbf{IDM}}(a_{t+1:t+k}\mid C_t,o_{t+1:t+k}).
\label{eq:joint}
\end{equation}
This factorization is a probabilistic decomposition, not a prescription for sequential sampling. The factor labels indicate predictive roles rather than distributions independently parameterized by individual FFN experts. Joint-WAM realizes the joint prediction with mutual visibility between future video and action tokens during denoising; thus, video-token representations can incorporate information from the evolving action tokens despite the VGM role name.

\textbf{IDM-WAM.} The visual factor is conditioned on the available observed context. Its predicted rollout subsequently supplies the visual condition for the inverse-dynamics factor:
\begin{equation}
p(o_{t+1:t+k},a_{t+1:t+k}\mid C_t)
=
p_{\mathbf{VGM}}(o_{t+1:t+k}\mid C_t)
p_{\mathbf{IDM}}(a_{t+1:t+k}\mid C_t,o_{t+1:t+k}).
\label{eq:idm}
\end{equation}
Here the order of the two factors also follows the generation procedure: the video rollout is obtained before the corresponding actions are inferred. The one-way attention mask prevents the video branch from depending on future action tokens.

\textbf{Action-Centric-WAM.} The VLA factor defines the action distribution from the available observed context, and the WM factor models future observations conditioned on those actions:
\begin{equation}
p(o_{t+1:t+k},a_{t+1:t+k}\mid C_t)
=
p_{\mathbf{VLA}}(a_{t+1:t+k}\mid C_t)
p_{\mathbf{WM}}(o_{t+1:t+k}\mid C_t,a_{t+1:t+k}).
\label{eq:gigaworld}
\end{equation}
The second factor adds an action-conditioned video objective during training. At inference, only the action distribution is sampled; no future video is generated, so WM is inactive.

\textbf{Action-Only-WAM.} With future cross-modal dependencies removed, the action and video factors share the observed context but do not condition on one another's future outputs:
\begin{equation}
p(o_{t+1:t+k},a_{t+1:t+k}\mid C_t)
=
p_{\mathbf{VLA}}(a_{t+1:t+k}\mid C_t)
p_{\mathbf{VGM}}(o_{t+1:t+k}\mid C_t).
\label{eq:fastwam}
\end{equation}
The VGM factor contributes auxiliary video supervision during training. At deployment, VGM remains active for the initial conditioning frame, while shared and VLA experts predict future action tokens without generating future video. This product form expresses the conditional independence imposed between the two future modalities.

\textbf{AC-WM.} The future action sequence appears entirely in the conditioning variables. The WM capability therefore defines a distribution over future observations only:
\begin{equation}
p(o_{t+1:t+k}\mid C_t,a_{t+1:t+k})
=
p_{\mathbf{WM}}(o_{t+1:t+k}\mid C_t,a_{t+1:t+k}).
\label{eq:acwm}
\end{equation}
Unlike the four WAM interfaces, this conditional contains no action-generation factor. The supplied action sequence remains fixed while the model predicts its visual consequences.

\subsection{Training and Inference Visibility Masks}
\label{sec:appendix-interface-masks}

\begin{figure}[H]
\centering
\includegraphics[width=0.82\linewidth]{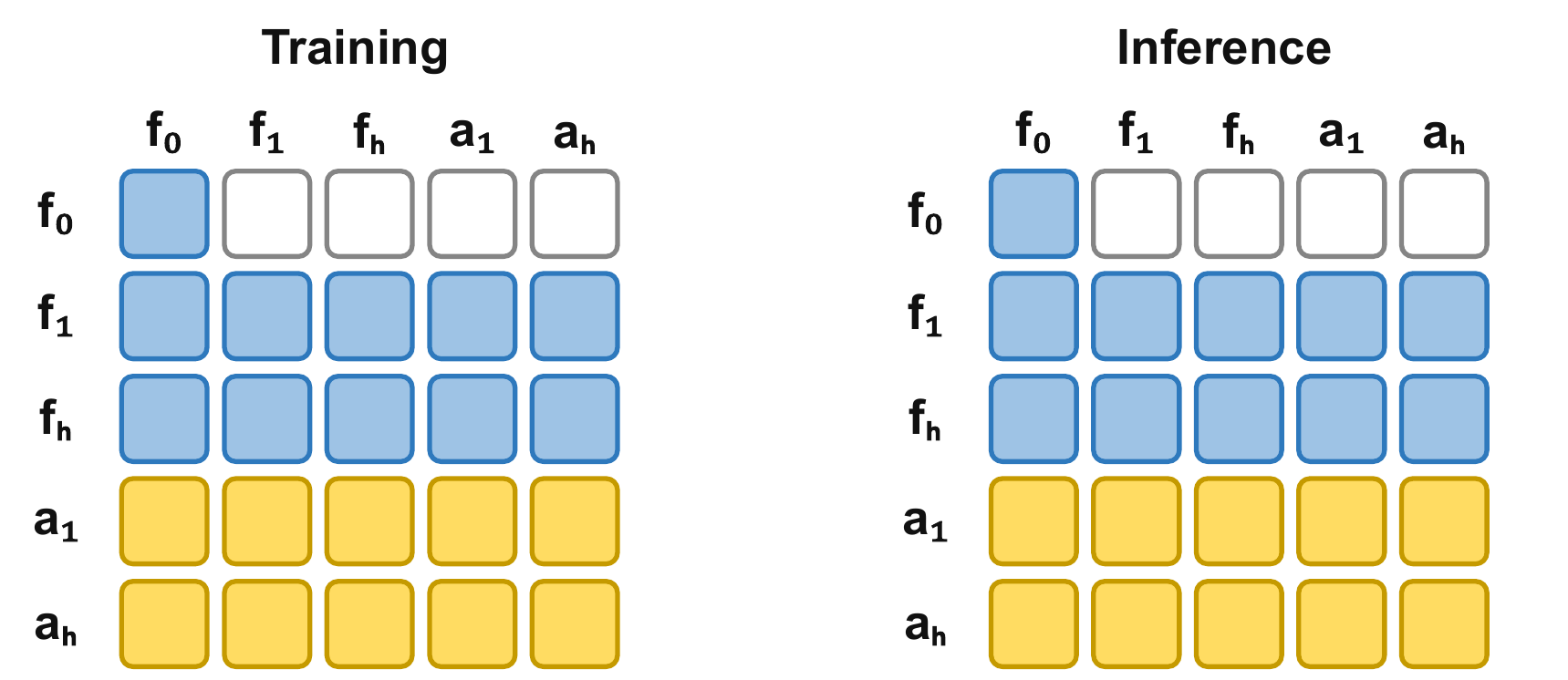}
\caption{Joint-WAM visibility masks. Future video and action tokens attend to both modalities, coupling their predictions during joint denoising. Training and inference use the same bidirectional visibility pattern.}
\label{fig:mask-joint}
\end{figure}

\begin{figure}[H]
\centering
\includegraphics[width=0.82\linewidth]{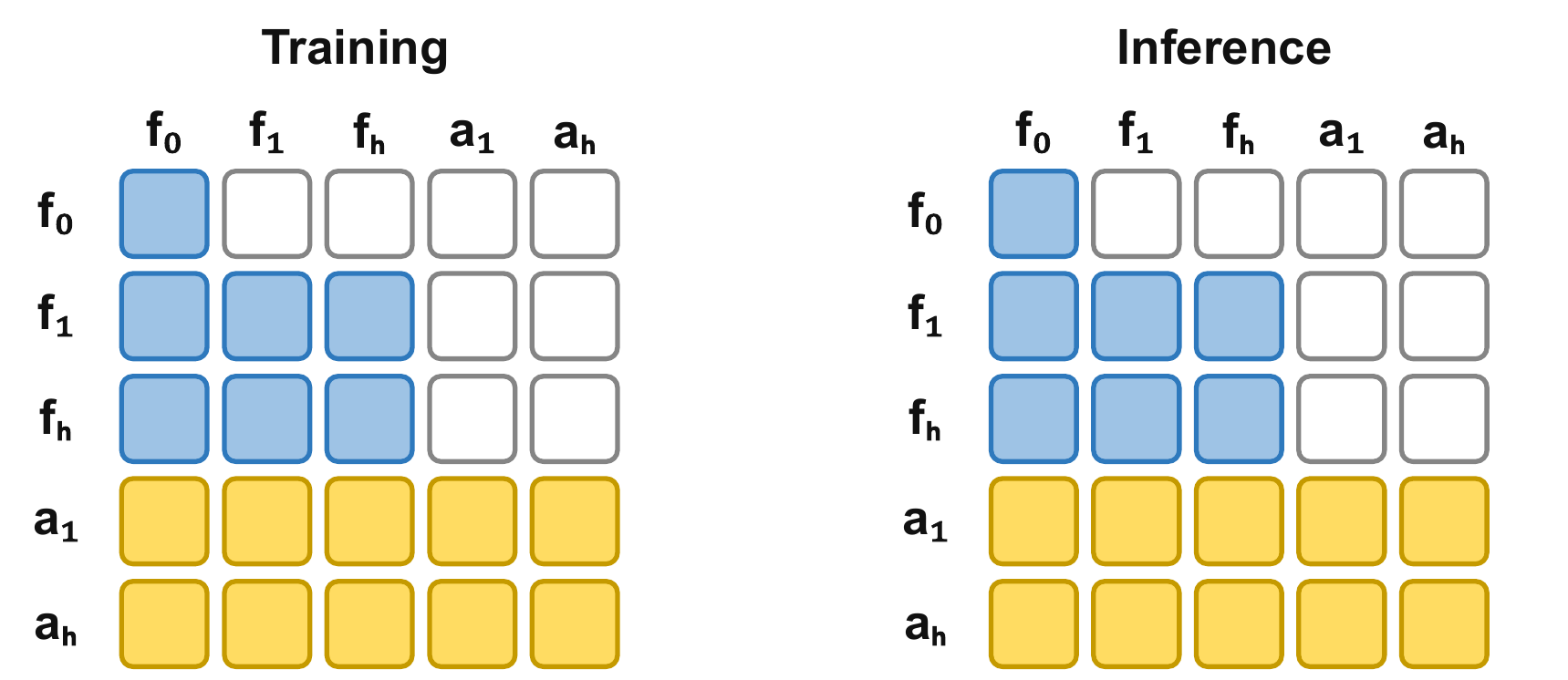}
\caption{IDM-WAM visibility masks. Action queries can access future video, while video queries cannot access future actions. At inference, video is generated first and then held fixed as the condition for action generation.}
\label{fig:mask-idm}
\end{figure}

\begin{figure}[H]
\centering
\includegraphics[width=0.82\linewidth]{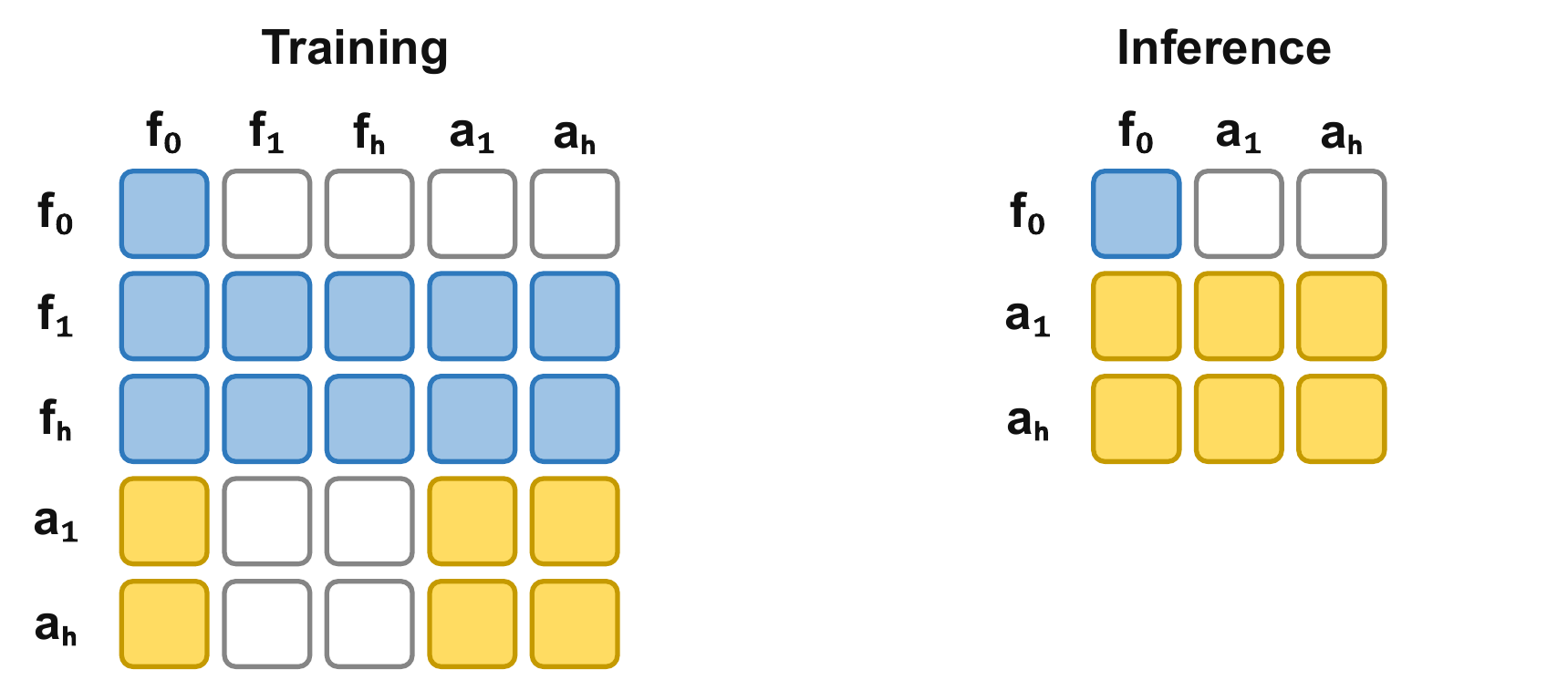}
\caption{Action-Centric-WAM visibility masks. During training, future video queries access actions, but action queries cannot access future video. At deployment, future video tokens are omitted and WM is inactive, while observed-context tokens remain available to action generation.}
\label{fig:mask-action-centric}
\end{figure}

\begin{figure}[H]
\centering
\includegraphics[width=0.82\linewidth]{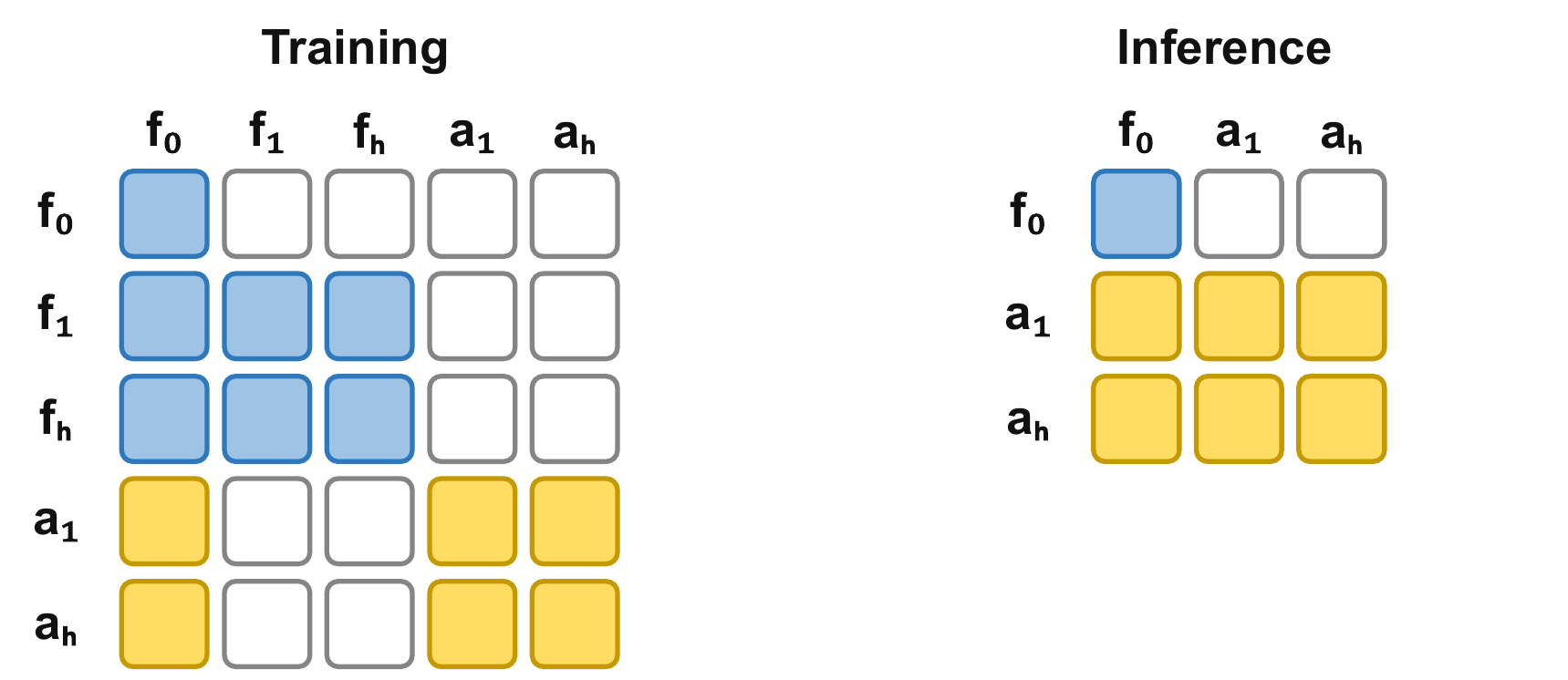}
\caption{Action-Only-WAM visibility masks. Neither future modality attends to the other during training, although both access the observed context. Inference retains only the context and action tokens, omitting auxiliary video prediction.}
\label{fig:mask-action-only}
\end{figure}

\begin{figure}[H]
\centering
\includegraphics[width=0.82\linewidth]{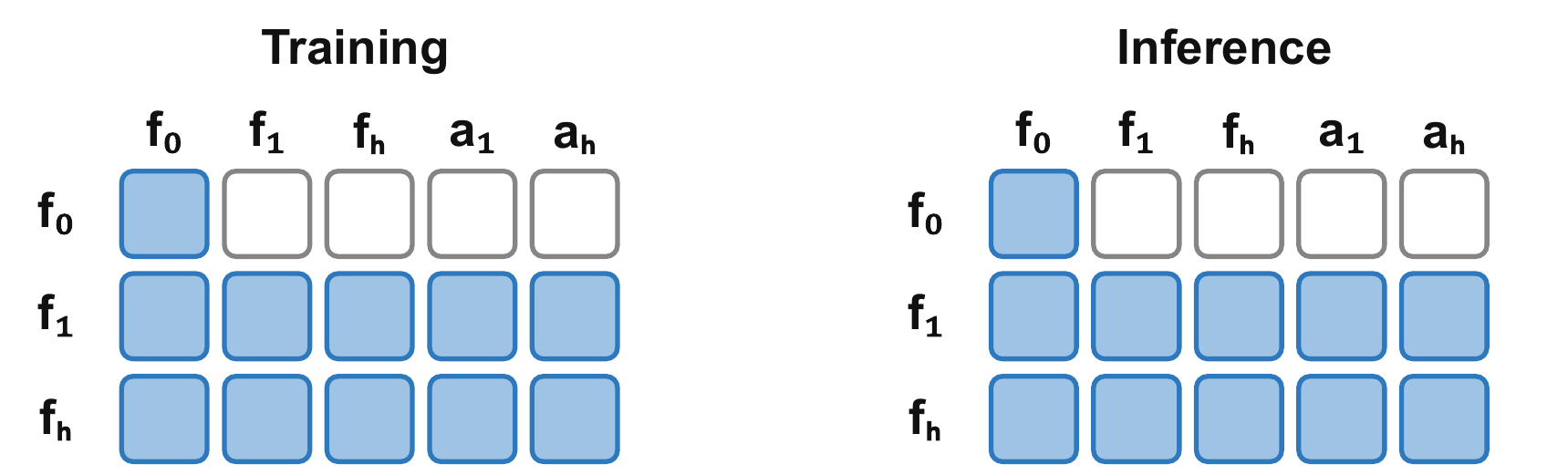}
\caption{AC-WM visibility masks. Future video queries attend to the supplied action sequence during both training and inference. Actions remain clean conditioning inputs, so the interface predicts video without generating actions.}
\label{fig:mask-acwm}
\end{figure}

Figures~\ref{fig:mask-joint}--\ref{fig:mask-acwm} visualize the attention dependencies of the five interfaces during training and inference. Rows denote query tokens and columns denote key tokens: a filled cell allows the row token to attend to the column token, whereas a white cell blocks that connection. Blue and yellow indicate video-query and action-query rows, respectively. Here $f_0$ denotes the observed visual context, $f_1$ and $f_h$ represent future video tokens, and $a_1$ and $a_h$ represent future action tokens. Other shared conditioning inputs are omitted for clarity.

The masks specify visibility rather than the denoising schedule. In particular, IDM-WAM uses sequential video-then-action inference despite retaining the same dependency pattern. AC-WM treats actions as fixed inputs; its diagrams therefore retain action columns but omit action-query rows. Their expansion into training windows and teacher-forcing copies is described in Appendix~\ref{sec:appendix-trajectory-organization}.

\subsection{Trajectory Organization and Teacher Forcing}
\label{sec:appendix-trajectory-organization}

The unified training configuration draws all five interfaces from the same trajectory-window format: 33 RGB frames, 96 action vectors, four robot states, and a language instruction. Video encoding produces nine temporal latent frames, comprising an initial condition and four successive blocks of two future latent frames. Each block is aligned with 24 actions and one state. The language instruction enters through cross-attention. AC-WM uses the aligned video and action data but does not include state tokens in its Transformer input.

Table~\ref{tab:training-trajectory-organization} distinguishes conditioning copies from prediction targets. Clean video copies provide observed history for teacher forcing; they are not additional prediction targets. For Joint-WAM and the full-window Action-Centric-WAM, Action-Only-WAM, and AC-WM paths, a target block accesses the clean video prefix preceding that block, together with the current-block inputs permitted by its interface mask. This prevents the clean copy of the target future video from becoming a direct input to its own prediction. IDM-WAM instead follows an interleaved video-then-action ordering: the current video condition precedes the action target, allowing action denoising to use that video while preventing the reverse dependence. Its video-condition augmentation is described in Appendix~\ref{sec:appendix-training-objectives}.

\begin{table}[t]
\centering
\caption{Training trajectory organization. All rows use the same four-block trajectory window. The initial video frame is conditioning only and excluded from video loss. Clean copies serve teacher forcing rather than additional supervision.}
\label{tab:training-trajectory-organization}
\small
\setlength{\tabcolsep}{4pt}
\renewcommand{\arraystretch}{1.12}
\begin{tabularx}{\linewidth}{lXX}
\toprule
\textbf{Interface} & \textbf{Conditioning copies and inputs} & \textbf{Denoising targets} \\
\midrule
Joint-WAM & Clean video history, current state, and language & Future video and actions, with joint current-block visibility \\
IDM-WAM & Video and action conditioning copies, states, and language under interleaved teacher forcing & Future video and actions, ordered video before action \\
Action-Centric-WAM & Clean video history, current state, and language & Future video and actions; video additionally reads current noisy actions \\
Action-Only-WAM & Clean video history, current state, and language & Future video and actions without current-block cross-modal attention \\
AC-WM & Clean video history, supplied clean actions, and language & Future video only \\
\bottomrule
\end{tabularx}
\end{table}

The visibility diagrams in Appendix~\ref{sec:appendix-interface-masks} summarize modality-level dependencies. The implementation expands these dependencies over temporal blocks and separate conditioning/target copies, allowing teacher-forced blocks to be processed in parallel without changing the prescribed information flow.

\subsection{Capability-Aligned Token Routing}
\label{sec:appendix-expert-activation}

\begin{figure}[H]
\centering
\includegraphics[width=\linewidth]{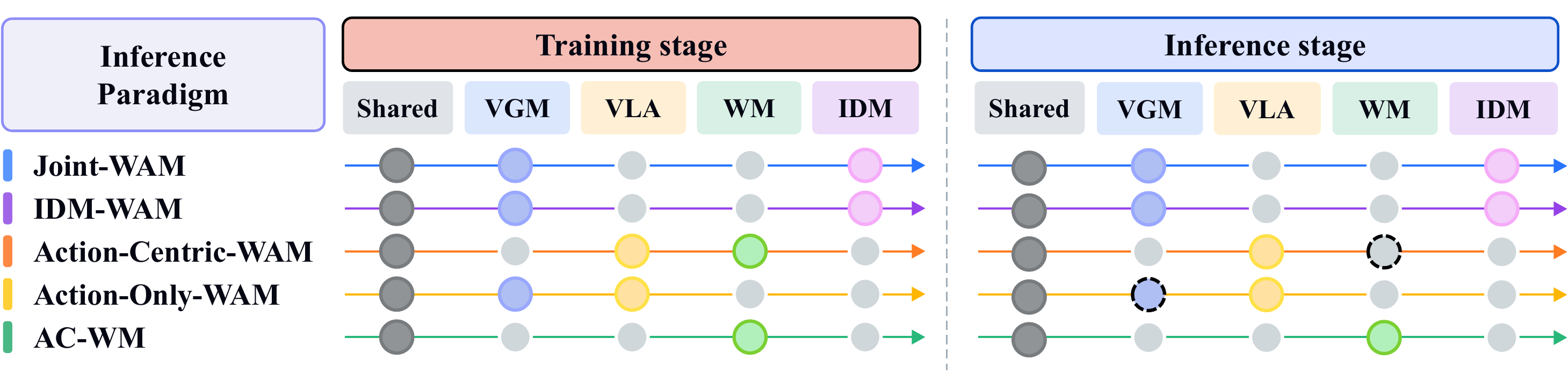}
\caption{Expert activation across five video-action interfaces. Shared experts are active for every interface, while atomic experts specialize according to the conditional pathway. In Action-Centric-WAM, the grey dashed WM circle indicates that the expert is inactive when future-video prediction is omitted. In Action-Only-WAM, the coloured dashed VGM circle indicates that the expert remains active for the initial conditioning frame, without predicting future video. For these two interfaces, future action tokens are processed only by shared and VLA experts.}
\label{fig:expert-activation}
\end{figure}

Figure~\ref{fig:expert-activation} summarizes the expert groups used by each interface in the two execution settings. The shared component is retained throughout, whereas the active atomic groups depend on which modalities are predicted and what they condition on. At inference, Action-Centric-WAM uses shared and VLA experts; WM is inactive because no future video is generated. Action-Only-WAM additionally retains VGM to process the initial conditioning-frame tokens, while future action tokens use shared and VLA experts. Its separate conditioning-frame expert path is retained even when future-video prediction is removed. This design preserves specialized visual conditioning within the interface's separation of future video and action targets; information leakage from future targets is prevented by the visibility mask, not by expert activation alone. The grey dashed WM circle denotes deactivation, whereas the coloured dashed VGM circle denotes retained conditioning-frame processing without future-video generation.

\paragraph{Allowed experts and learned mixture weights.}
CATR combines a prescribed expert eligibility set with learned mixing weights. Each MoE feed-forward sublayer contains one shared expert and four atomic experts, VGM, VLA, WM, and IDM. Table~\ref{tab:catr-token-experts} gives the eligible experts for future prediction tokens. In the teacher-forced layout, clean history copies and state tokens use the shared expert alone; clean action conditions in AC-WM also use only the shared expert. These clean history copies are distinct from the initial conditioning-frame (seed) group: Action-Only-WAM routes that group through shared and VGM experts during both training and inference, whereas Action-Centric-WAM uses shared alone. In cached inference, clean history is written through shared alone; VGM participates in the Action-Only-WAM seed pass, and subsequent action queries use shared and VLA experts while reading the cached context. Thus, VGM does not process every historical frame, and WM is not introduced into the Action-Centric-WAM inference path. The table concerns future target groups rather than conditioning-frame or clean-history copies.

\begin{table}[H]
\centering
\caption{Eligible experts for future target groups during training. Weights are learned within each listed set; experts outside the set do not contribute to that group's output.}
\label{tab:catr-token-experts}
\small
\setlength{\tabcolsep}{7pt}
\begin{tabular}{lll}
\toprule
\textbf{Interface} & \textbf{Future video targets} & \textbf{Action targets} \\
\midrule
Joint-WAM / IDM-WAM & Shared + VGM & Shared + IDM \\
Action-Centric-WAM & Shared + WM & Shared + VLA \\
Action-Only-WAM & Shared + VGM & Shared + VLA \\
AC-WM & Shared + WM & None (clean conditions) \\
\bottomrule
\end{tabular}
\end{table}

For a routed token group $g$ (a prediction-target or conditioning group), let $\mathcal{E}_g$ be its eligible expert set and $\mathcal{R}_g$ the token indices pooled for routing. At a given Transformer layer, the router operates on hidden states $h_j$ after attention:
\begin{equation}
    \bar h_g=\frac{1}{|\mathcal{R}_g|}\sum_{j\in\mathcal{R}_g}h_j,
    \qquad
    \alpha_{g,e}=
    \begin{cases}
    \displaystyle\frac{\exp((W_r\bar h_g)_e/\tau)}
    {\sum_{e'\in\mathcal{E}_g}\exp((W_r\bar h_g)_{e'}/\tau)},
    & e\in\mathcal{E}_g,\\[6pt]
    0, & e\notin\mathcal{E}_g.
    \end{cases}
    \label{eq:appendix-catr-weights}
\end{equation}
Here, $W_r$ is the learned router matrix and $\tau$ is its temperature. For each token $j$ in group $g$, the feed-forward output is
\begin{equation}
    \operatorname{MoE}(h_j)
    =\sum_{e\in\mathcal{E}_g}\alpha_{g,e}f_e(h_j),
    \label{eq:appendix-catr-mixture}
\end{equation}
where $f_e$ is expert $e$. Thus, tokens within a group share mixture weights but retain their individual expert outputs. Shared-only groups have a fixed weight of one and need no learned mixture.

\begin{table}[H]
\centering
\caption{Expert sets and mixing rules for conditioning and action groups in the two action-only inference pathways. Shared-only groups have fixed weight one; learned mixtures use Equations~\ref{eq:appendix-catr-weights} and~\ref{eq:appendix-catr-mixture}.}
\label{tab:catr-conditioning-groups}
\footnotesize
\setlength{\tabcolsep}{4pt}
\begin{tabularx}{\linewidth}{@{}Xlll@{}}
\toprule
\textbf{Token group} & \textbf{Action-Centric} & \textbf{Action-Only} & \textbf{Mixing} \\
\midrule
Clean history copies & Shared & Shared & Fixed \\
Initial conditioning frame (seed) & Shared & Shared + VGM & Fixed / learned \\
Future action targets & Shared + VLA & Shared + VLA & Learned \\
State tokens & Shared & Shared & Fixed \\
\bottomrule
\end{tabularx}
\end{table}

For the Action-Only-WAM seed group, $\mathcal{R}_g$ contains that group's frame-token indices: their post-attention hidden states are mean-pooled, projected by $W_r$, and normalized over shared and VGM experts. The resulting weights are shared within the group, while each token retains its own expert output. Action-Centric-WAM uses only shared for this group. In the cached action-query pass, action-group weights are similarly obtained by pooling the current action tokens, whose hidden states already incorporate the permitted cached context.

\paragraph{Caching and interface changes.}
Clean-history and seed passes store attention keys and values in the interface's KV cache for subsequent action queries; the expert-mixture weights are computed for each routed group rather than reused as cache entries. In the real-robot PAIR server, an interface change clears the model's runtime caches and resets window tracking. The newly selected interface is seeded from the latest frame of the current observation, rebuilding its cache through its own computation path; KV entries from the previous interface are not transferred. If the interface remains unchanged, normal within-interface caching continues, subject to its window resets.

\paragraph{Pooling scope and interface selection.}
Joint-WAM pools the current block's video, action, and state representations, then applies separate eligibility masks for video and action groups. IDM-WAM pools each video or action target group separately. The full-window Action-Centric-WAM, Action-Only-WAM, and AC-WM paths likewise pool each predicted modality within its block. These representations already incorporate the context allowed by attention; the router does not pool the entire trajectory indiscriminately. CATR therefore specifies expert computation inside the selected interface, whereas PAIR selects the action-generating interface at deployment. The two routing mechanisms operate at different levels.

\section{PAIR: Training and Deployment}
\label{sec:appendix-pair}

\subsection{MAE-Based Training Targets}
\label{sec:appendix-pair-targets}

The OOD adaptation datasets comprise human-collected demonstrations of the same three real-world tasks: Folding Clothes, Folding Box, and Tabletop Cleaning. Each episode records camera observations and the corresponding robot actions throughout task execution. These demonstrations are separate from the data used to train the backbone: the task goals are shared, but the object instances and scene conditions differ, including lighting and object appearance such as color. Task-specific shifts include garment colors and sizes, box types, and cluttered backgrounds, as defined in Section~\ref{sec:exp-setup}.

For each target task, we randomly select 500 episodes from its OOD dataset and construct candidate chunk scenarios using a sliding window of 24 consecutive action steps with stride one action step, aligned with the execution chunk length. We then randomly sample 2,000 chunk scenarios from this candidate pool. Each scenario is evaluated by all four frozen WAM interfaces, producing 8,000 interface-specific action predictions per task. The reference action chunk is the recorded action sequence from the corresponding window in the OOD dataset, not an action sequence generated by any of the four interfaces. We compare all four predictions with this same reference to obtain four MAEs and identify the lowest-error interface for each scenario. These errors provide the routing supervision, with soft targets and the near-best efficiency preference defined below. Thus, each task contributes 2,000 labeled chunk scenarios, not 8,000 independent training windows; all four interfaces are evaluated on every scenario rather than sampled as separate data sources. Using this 24-step, stride-one counting convention, sliding-window construction across the three task datasets yields up to 21,867,390 candidate windows in total. These overlapping windows are candidate chunk scenarios, not independent demonstrations. The 6,000 selected scenarios across the three tasks constitute approximately 0.0274\% of this candidate pool (less than one thousandth). This fraction measures window sampling, not a reduction in episode acquisition requirements. PAIR is trained offline with the UniW(A)M backbone frozen. Final OOD success rates are measured in new physical robot trials rather than on windows held out from these episodes; the deployment environments do not exactly reproduce the dataset configurations, as detailed in Appendix~\ref{sec:appendix-benchmarks}.

For an action chunk of $k$ steps, the frozen backbone generates candidate action sequences $\hat{a}_{t+1:t+k}^{(i)}$ for the four WAM interfaces:
\begin{equation}
\mathcal{I}_{\mathrm{WAM}}
=
\left\{
\mathrm{Joint\text{-}WAM},
\mathrm{IDM\text{-}WAM},
\mathrm{Action\text{-}Centric\text{-}WAM},
\mathrm{Action\text{-}Only\text{-}WAM}
\right\}.
\label{eq:wam-interface-set}
\end{equation}
Given the ground-truth sequence $a_{t+1:t+k}^{*}$, the action error for each interface is
\begin{equation}
\mathcal{L}_{\mathrm{MAE}}^{(i)}
=
\frac{1}{k d_a}
\sum_{s=t+1}^{t+k}
\sum_{d=1}^{d_a}
\left|
\hat{a}_{s,d}^{(i)} - a_{s,d}^{*}
\right|,
\qquad
i \in \mathcal{I}_{\mathrm{WAM}},
\label{eq:router-mae}
\end{equation}
where $d_a$ is the action dimensionality. We normalize errors within each chunk scenario:
\begin{equation}
\Delta_t^{(i)}
=
\frac{
\mathcal{L}_{\mathrm{MAE}}^{(i)}
-
\min\limits_{j\in\mathcal{I}_{\mathrm{WAM}}}
\mathcal{L}_{\mathrm{MAE}}^{(j)}
}{
\max\!\left\{
\max\limits_{j\in\mathcal{I}_{\mathrm{WAM}}}
\mathcal{L}_{\mathrm{MAE}}^{(j)}
-
\min\limits_{j\in\mathcal{I}_{\mathrm{WAM}}}
\mathcal{L}_{\mathrm{MAE}}^{(j)}
,\epsilon\right\}
},
\label{eq:router-error-normalization}
\end{equation}
where $\epsilon=10^{-12}$ prevents division by zero. Equation~\ref{eq:router-soft-target} maps these normalized errors to the soft target $\tilde y_t^{(i)}$. Let $e_t^{(i)}=\mathcal{L}_{\mathrm{MAE}}^{(i)}$ and $e_t^*=\min_i e_t^{(i)}$. We define the near-best candidate set by
\begin{equation}
    \mathcal{N}_t=\left\{i\in\mathcal{I}_{\mathrm{WAM}}:
    e_t^{(i)}-e_t^*\leq\delta e_t^*\right\},
    \qquad \delta=0.01.
    \label{eq:router-relative-tie}
\end{equation}
For a positive minimum error, this includes every interface whose MAE is within 1\% of the minimum. If $e_t^*=0$, only zero-error interfaces qualify, avoiding division by zero. Let $\mathcal{F}$ contain Action-Centric-WAM and Action-Only-WAM. If $\mathcal{N}_t\cap\mathcal{F}$ is nonempty, we select its lowest-latency member $i_t^{\mathrm{fast}}$ and assign a one-hot training target:
\begin{equation}
    y_t^{(i)}=
    \begin{cases}
        \mathbf{1}\{i=i_t^{\mathrm{fast}}\}, & \mathcal{N}_t\cap\mathcal{F}\neq\emptyset,\\
        \tilde y_t^{(i)}, & \mathcal{N}_t\cap\mathcal{F}=\emptyset.
    \end{cases}
    \label{eq:router-efficiency-target}
\end{equation}
When both fast interfaces qualify, Action-Centric-WAM is selected according to the measured latency ordering in Section~\ref{sec:exp-pareto}. This preference is restricted to near-best candidates and does not assign probability to an ineligible fast interface. The rule constructs offline training labels; deployment uses the learned router without ground-truth actions or online MAE comparisons.

\subsection{Router Inputs and Chunk-Level Selection}
\label{sec:appendix-pair-inference}

PAIR uses contextual features from the frozen UniW(A)M backbone:
\begin{equation}
z_t
=
\Psi_{\theta}
\left(
F_t^{\mathrm{video}},
F_t^{\mathrm{CLIP}},
F_t^{\mathrm{state}},
F_t^{\mathrm{text}}, c_t
\right),
\label{eq:router-features}
\end{equation}
where the four feature inputs are video, CLIP-derived semantic, robot-state, and language features, respectively, $c_t$ is a one-hot encoding of the chunk position within the current window, and $\theta$ denotes the frozen backbone parameters. To obtain the video features, the current visible observation prefix is encoded and passed through all frozen DiT blocks with only the shared FFN expert enabled; the resulting video tokens are mean-pooled. This feature-extraction pass is independent of the interface to be selected and does not populate that interface's KV cache. Its pooled video representation is combined with the CLIP, state, language, and chunk-position features for routing. PAIR uses eight independently trained classification heads. Their softmax probabilities are averaged to obtain
\begin{equation}
q_{\phi}(i\mid z_t)
=
\frac{1}{8}\sum_{m=1}^{8}\operatorname{softmax}_{i}
\left(
g_{\phi_m}(z_t)
\right),
\qquad
i \in \mathcal{I}_{\mathrm{WAM}},
\label{eq:adaptive-router}
\end{equation}
where $\phi=\{\phi_m\}_{m=1}^{8}$ and each head is trained with the loss in Equation~\ref{eq:router-loss}. The PAIR results in Table~\ref{tab:ood-real} and Figure~\ref{fig:latency-tradeoff} use this same eight-head ensemble. At deployment, PAIR selects
\begin{equation}
i_t
=
\operatorname*{arg\,max}_{i\in\mathcal{I}_{\mathrm{WAM}}}
q_{\phi}(i\mid z_t).
\label{eq:router-inference}
\end{equation}
The selected WAM interface generates the next action chunk. After its execution, the observation history is updated before PAIR selects an interface for the next chunk. As detailed in Appendix~\ref{sec:appendix-expert-activation}, an interface change clears runtime caches and resets window tracking; the latest observed frame then seeds the selected interface's own cache. When the interface is unchanged, normal within-interface caching continues.

\paragraph{Router timing.} The eight heads are evaluated in parallel; their combined prediction and probability aggregation take approximately 2~ms in a separate measurement. To account for shared-expert feature computation and the remaining routing operations, we benchmark the first chunk of each window on a single NVIDIA H20 using checkpoint-150000 and an ensemble of eight router heads. After one warm-up window, we average five windows using synthetic 640$\times$480 camera inputs and CUDA synchronization at each timing boundary. Subtracting the separately timed observation and language encoding stage from the complete observation-to-selection call gives 62.7~ms on average, rounded to 63~ms. This accounting includes input preparation, shared-expert feature computation, classification, and interface selection, but excludes observation and language encoding, subsequent action generation, cache reconstruction, and multi-GPU control communication. It characterizes the first-chunk setting rather than an average over chunk positions. The synthetic inputs are used to measure computation time, not routing accuracy.

\subsection{Interface Selection across Execution Phases}
\label{sec:appendix-pair-selection}

\begin{figure}[H]
\centering
\includegraphics[width=\linewidth]{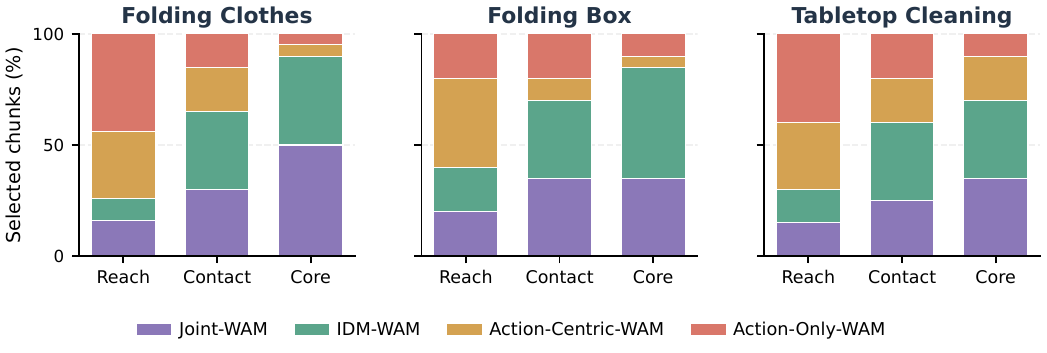}
\caption{Phase-conditioned PAIR interface selection across three real-world tasks. Each panel reports 100\%-stacked proportions for the Reach, Contact, and Core stages, revealing how the router shifts between fast action pathways and video-aware WAMs as interaction demands change.}
\label{fig:selector}
\end{figure}

Figure~\ref{fig:selector} reports phase-conditioned interface selection across the three real-world tasks. We divide execution into Reach, involving motion for approaching the target; Contact, involving contact-based manipulation; and Core, involving fine-grained manipulation, and measure each interface's selection frequency within each phase. During Reach, Action-Only-WAM and Action-Centric-WAM together account for 60\%--75\% of selections, favoring fast action pathways for low-interaction motions. During Contact, the combined share of Joint-WAM and IDM-WAM rises to approximately 60\%--70\%, consistent with a preference for video-conditioned geometric guidance. During Core, these two interfaces account for 70\%--90\% of selections, indicating greater use of visual foresight during fine manipulation and complex physical interactions. This phase-dependent pattern is consistent with balancing fast action execution and foresight-rich control as interaction demands change, and offers an interpretation of the OOD gains reported in the main text. Examples of critical manipulation steps are discussed in Appendix~\ref{sec:appendix-pair-case-study}.

\section{Benchmarks and Evaluation Protocols}
\label{sec:appendix-benchmarks}

We describe the benchmarks and evaluation protocols summarized in Section~\ref{sec:exp-setup}. LIBERO, RoboTwin 2.0, and the real-world tasks evaluate action generation, while WorldArena evaluates action-conditioned future-world prediction.

\paragraph{LIBERO.}

The benchmark comprises four official suites, including LIBERO-Spatial, LIBERO-Object, LIBERO-Goal, and LIBERO-Long, each containing 10 tasks with 50 demonstrations per task. We follow the standard public benchmarking protocol for training configurations and evaluation trial counts. We report task success rates for each suite alongside the overall average.

\paragraph{RoboTwin 2.0.}

RoboTwin 2.0 evaluates dual-arm manipulation across 50 tasks. Following the official benchmark setting~\citep{robotwin-leaderboard-2026}, we jointly train a single policy on 50 clean demonstrations per task (2,500 demonstrations in total). We evaluate the same policy for 100 trials per task under clean (Clean-to-Clean, C2C) and randomized (Clean-to-Randomized, C2R) conditions, reporting the mean success rate across the 50 tasks separately for each setting. Baselines come from the official leaderboard snapshot updated on September 24, 2026. The $\pi_0$ entry uses a separately fine-tuned checkpoint per task; the other listed baselines use co-training, which jointly trains one policy across all tasks.

\paragraph{Real-World Evaluation.}

Real-robot experiments are conducted on an AgileX dual-arm platform across three representative manipulation tasks: \textit{Folding Clothes}, \textit{Folding Box}, and \textit{Tabletop Cleaning}. Each task is evaluated over 20 trials. All main comparisons and ablation studies share these three tasks, with task success rate serving as the primary evaluation metric. For PAIR, offline adaptation uses recorded OOD dataset episodes as described in Appendix~\ref{sec:appendix-pair-targets}, whereas final OOD evaluation consists of separately conducted real-robot deployments under environment configurations that do not exactly match those in the dataset. This evaluates transfer from offline target-domain data to physical execution; it is not a random window-level train/test split of the same recorded episodes.

\paragraph{Efficiency metrics.}
\label{sec:appendix-efficiency-metrics}
All interfaces use a configured denoising step count of 16. Joint-WAM jointly denoises video and actions for 16 steps, while Action-Centric-WAM and Action-Only-WAM each denoise actions for 16 steps. IDM-WAM performs 16 video-denoising steps followed by 16 action-denoising steps. CFG is set to 5 for Joint-WAM and for IDM-WAM's video stage; IDM-WAM's action stage uses no CFG, and both fast interfaces use a guidance scale of 1. The standalone interface latency in Figure~\ref{fig:latency-tradeoff}(b) is the elapsed time for a complete action-chunk inference call divided by the configured step count of 16. For IDM-WAM, this denominator is 16 rather than the 32 iterations summed across its two stages. Multiplying each reported normalized latency by 16 recovers the corresponding full chunk inference time. Initial conditioning and cache construction performed within that call are included and amortized over the steps; the metric is not a direct measurement of a steady-state denoising iteration. PAIR routing is timed separately as described in Appendix~\ref{sec:appendix-pair-inference}. The task-level completion time in Figure~\ref{fig:latency-tradeoff}(a) is averaged over successful trials only. Failed and timed-out trials are excluded from this time average but remain in the success-rate denominator.

\paragraph{WorldArena.}

To evaluate future world prediction, UniW(A)M operates as an action-conditioned world model following the standard WorldArena evaluation protocol. We report metrics across visual quality, motion quality, content consistency, physical adherence, 3D accuracy, and controllability, comparing against established action-conditioned video prediction baselines under the same benchmark protocol. This evaluation verifies whether unified multi-task training preserves action-conditioned world modeling capabilities.

\subsection{Uncertainty in Real-World Success Rates}
\label{sec:appendix-uncertainty}

We report per-task success counts and two-sided 95\% Wilson score intervals without continuity correction~\citep{wilson1927probable} for the existing real-robot evaluations. Each method--task setting consists of $n=20$ trials, so the reported task success rate determines the integer count $k$. These calculations summarize the existing trials and do not introduce additional robot evaluations. With $\hat p=k/n$ and $z=1.959964$, the interval is
\begin{equation}
\frac{\hat p+z^2/(2n)\ \pm\ z\sqrt{\hat p(1-\hat p)/n+z^2/(4n^2)}}{1+z^2/n}.
\label{eq:wilson-interval}
\end{equation}
The calculation assumes approximately independent Bernoulli outcomes with a common success probability within each method--task setting. The intervals quantify finite-trial uncertainty conditional on the evaluated model and task setting; they do not capture variation across training runs. We report marginal intervals for individual task success rates, not simultaneous confidence bounds or tests of differences between methods. We do not pool the three tasks as 60 trials with a common success probability or assign intervals to cross-task averages without the corresponding trial structure.

Table~\ref{tab:real-world-wilson} supplements the standard and OOD evaluations in Tables~\ref{tab:real} and~\ref{tab:ood-real}. Table~\ref{tab:ablation-wilson} covers the dense, sample-level routing, and four-WAM-only ablations in Table~\ref{tab:ablate-mix}; the full five-paradigm CATR results reuse the standard UniW(A)M rows in Table~\ref{tab:real-world-wilson}. One additional success changes a task's observed rate by five percentage points, so small differences in point estimates should not be interpreted as stable performance rankings.

\begin{table}[htbp]
\centering
\caption{Per-task uncertainty for standard and OOD real-world evaluations. Each entry gives successes out of 20 trials, followed by the 95\% Wilson interval in percent. Standard and OOD rows correspond to separate evaluation settings.}
\label{tab:real-world-wilson}
\scriptsize
\setlength{\tabcolsep}{4pt}
\renewcommand{\arraystretch}{1.15}
\begin{tabular}{lccc}
\toprule
\textbf{Method / interface} & \textbf{Folding Clothes} & \textbf{Folding Box} & \textbf{Tabletop Cleaning} \\
\midrule
\rowcolor{gray!15}
\multicolumn{4}{c}{\textit{Standard evaluation: baselines}} \\
$\pi_{0.5}$~\citep{250416054} & 12/20; [38.7, 78.1] & 13/20; [43.3, 81.9] & 14/20; [48.1, 85.5] \\
DreamZero~\citep{260215922} & 9/20; [25.8, 65.8] & 11/20; [34.2, 74.2] & 12/20; [38.7, 78.1] \\
WorldScape Policy 2.0~\citep{260718840} & 15/20; [53.1, 88.8] & 15/20; [53.1, 88.8] & 16/20; [58.4, 91.9] \\
\midrule
\rowcolor{gray!15}
\multicolumn{4}{c}{\textit{Standard evaluation: UniW(A)M with CATR}} \\
Joint-WAM & 18/20; [69.9, 97.2] & 16/20; [58.4, 91.9] & 18/20; [69.9, 97.2] \\
IDM-WAM & 18/20; [69.9, 97.2] & 17/20; [64.0, 94.8] & 18/20; [69.9, 97.2] \\
Action-Centric-WAM & 17/20; [64.0, 94.8] & 15/20; [53.1, 88.8] & 17/20; [64.0, 94.8] \\
Action-Only-WAM & 17/20; [64.0, 94.8] & 14/20; [48.1, 85.5] & 17/20; [64.0, 94.8] \\
\midrule
\rowcolor{gray!15}
\multicolumn{4}{c}{\textit{OOD evaluation}} \\
Joint-WAM & 13/20; [43.3, 81.9] & 11/20; [34.2, 74.2] & 14/20; [48.1, 85.5] \\
IDM-WAM & 13/20; [43.3, 81.9] & 10/20; [29.9, 70.1] & 15/20; [53.1, 88.8] \\
Action-Centric-WAM & 10/20; [29.9, 70.1] & 8/20; [21.9, 61.3] & 11/20; [34.2, 74.2] \\
Action-Only-WAM & 10/20; [29.9, 70.1] & 9/20; [25.8, 65.8] & 11/20; [34.2, 74.2] \\
UniW(A)M + PAIR & 15/20; [53.1, 88.8] & 14/20; [48.1, 85.5] & 17/20; [64.0, 94.8] \\
\bottomrule
\end{tabular}
\end{table}

\begin{table}[htbp]
\centering
\caption{Per-task uncertainty for the dense, sample-level routing, and four-WAM-only ablations. Entries give successes out of 20 trials and 95\% Wilson intervals in percent. Counts and intervals for the full five-paradigm CATR model appear in the standard UniW(A)M block of Table~\ref{tab:real-world-wilson}.}
\label{tab:ablation-wilson}
\scriptsize
\setlength{\tabcolsep}{4pt}
\renewcommand{\arraystretch}{1.15}
\begin{tabular}{lccc}
\toprule
\textbf{Method / interface} & \textbf{Folding Clothes} & \textbf{Folding Box} & \textbf{Tabletop Cleaning} \\
\midrule
\rowcolor{gray!15}
\multicolumn{4}{c}{\textit{Single DiT (Dense)}} \\
Joint-WAM & 2/20; [2.8, 30.1] & 2/20; [2.8, 30.1] & 4/20; [8.1, 41.6] \\
IDM-WAM & 4/20; [8.1, 41.6] & 0/20; [0.0, 16.1] & 6/20; [14.5, 51.9] \\
Action-Centric-WAM & 0/20; [0.0, 16.1] & 0/20; [0.0, 16.1] & 2/20; [2.8, 30.1] \\
Action-Only-WAM & 0/20; [0.0, 16.1] & 0/20; [0.0, 16.1] & 1/20; [0.9, 23.6] \\
\midrule
\rowcolor{gray!15}
\multicolumn{4}{c}{\textit{MoE with Sample-Level Routing}} \\
Joint-WAM & 10/20; [29.9, 70.1] & 9/20; [25.8, 65.8] & 12/20; [38.7, 78.1] \\
IDM-WAM & 16/20; [58.4, 91.9] & 6/20; [14.5, 51.9] & 14/20; [48.1, 85.5] \\
Action-Centric-WAM & 2/20; [2.8, 30.1] & 0/20; [0.0, 16.1] & 4/20; [8.1, 41.6] \\
Action-Only-WAM & 2/20; [2.8, 30.1] & 0/20; [0.0, 16.1] & 4/20; [8.1, 41.6] \\
\midrule
\rowcolor{gray!15}
\multicolumn{4}{c}{\textit{UniW(A)M with CATR: Four WAMs Only (w/o AC-WM)}} \\
Joint-WAM & 15/20; [53.1, 88.8] & 12/20; [38.7, 78.1] & 16/20; [58.4, 91.9] \\
IDM-WAM & 16/20; [58.4, 91.9] & 12/20; [38.7, 78.1] & 16/20; [58.4, 91.9] \\
Action-Centric-WAM & 11/20; [34.2, 74.2] & 6/20; [14.5, 51.9] & 13/20; [43.3, 81.9] \\
Action-Only-WAM & 10/20; [29.9, 70.1] & 2/20; [2.8, 30.1] & 10/20; [29.9, 70.1] \\
\bottomrule
\end{tabular}
\end{table}

\section{Real-World Task Visualizations}
\label{sec:appendix-real-world-visualizations}

\subsection{Task Sequences across WAM Interfaces}
\label{sec:appendix-real-world-sequences}

\begin{figure}[H]
\centering
\includegraphics[width=\linewidth]{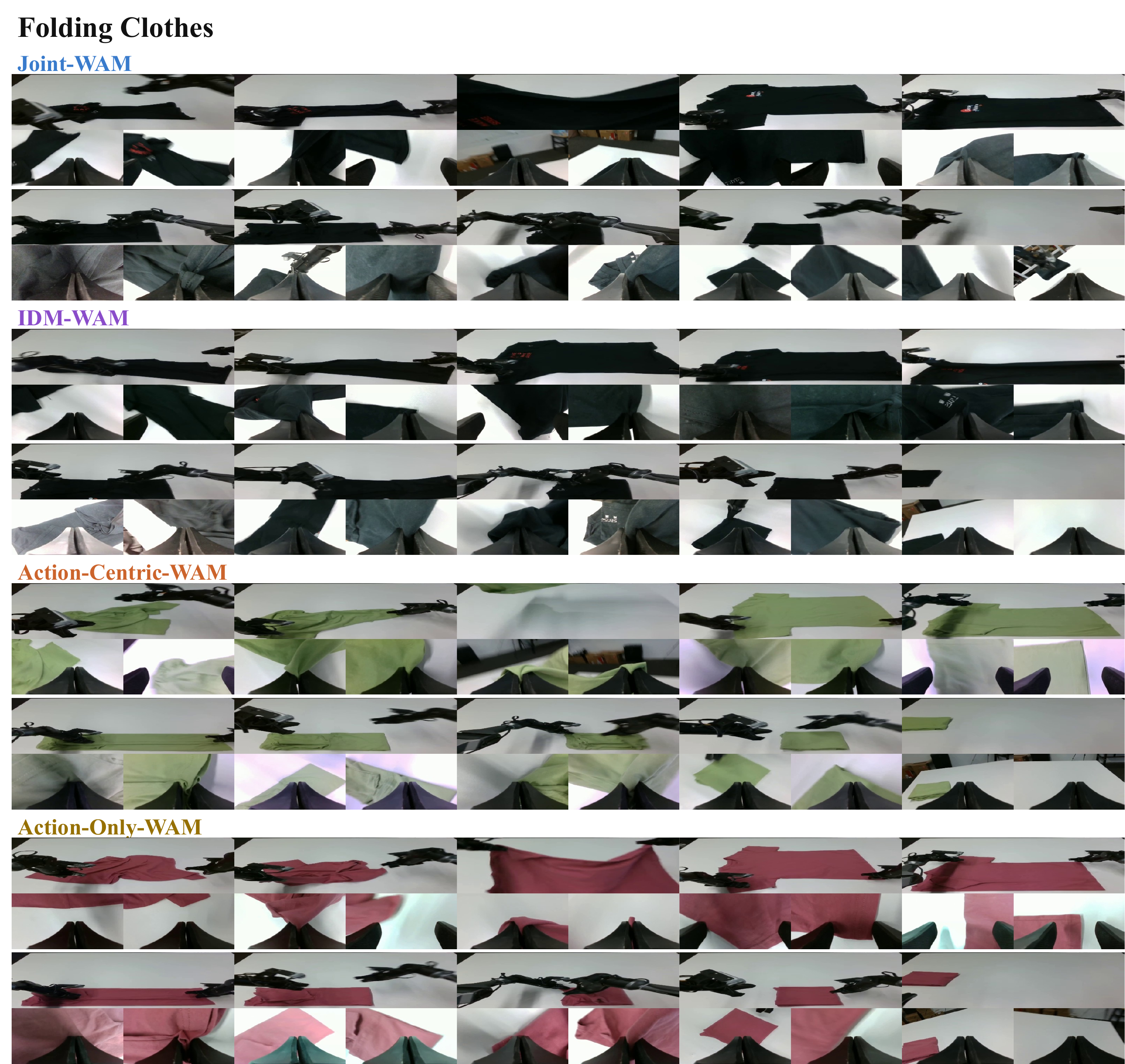}
\caption{Real-world visualizations of Folding Clothes for the four WAM interfaces. The sequences illustrate garment grasping, repositioning, and folding through successive stages of manipulation.}
\label{fig:real-world-clothes}
\end{figure}

\begin{figure}[H]
\centering
\includegraphics[width=\linewidth]{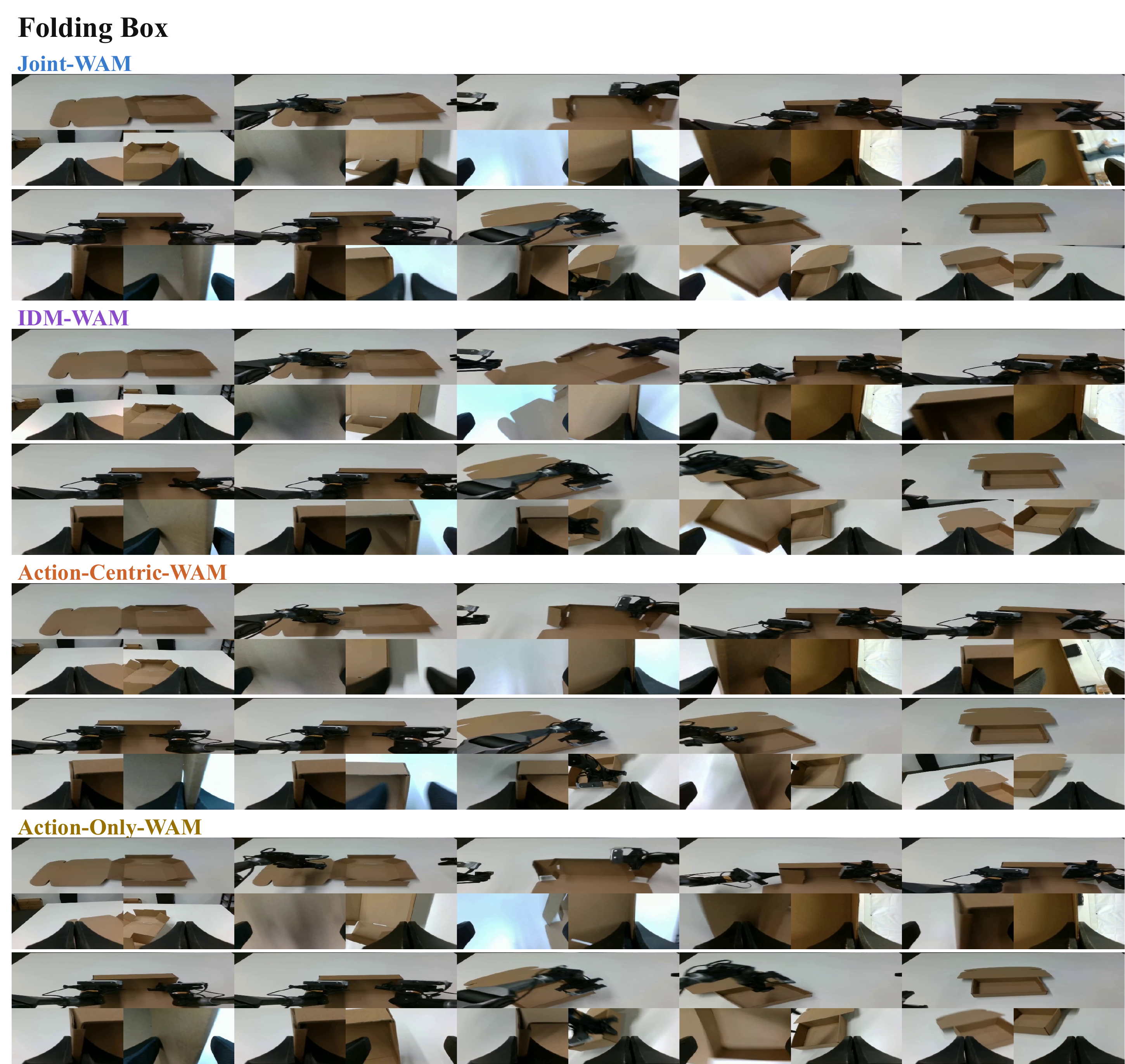}
\caption{Real-world visualizations of Folding Box for the four WAM interfaces. Selected frames show the progression from a flat cardboard box to folding its walls and manipulating its flaps.}
\label{fig:real-world-box}
\end{figure}

\begin{figure}[H]
\centering
\includegraphics[width=\linewidth]{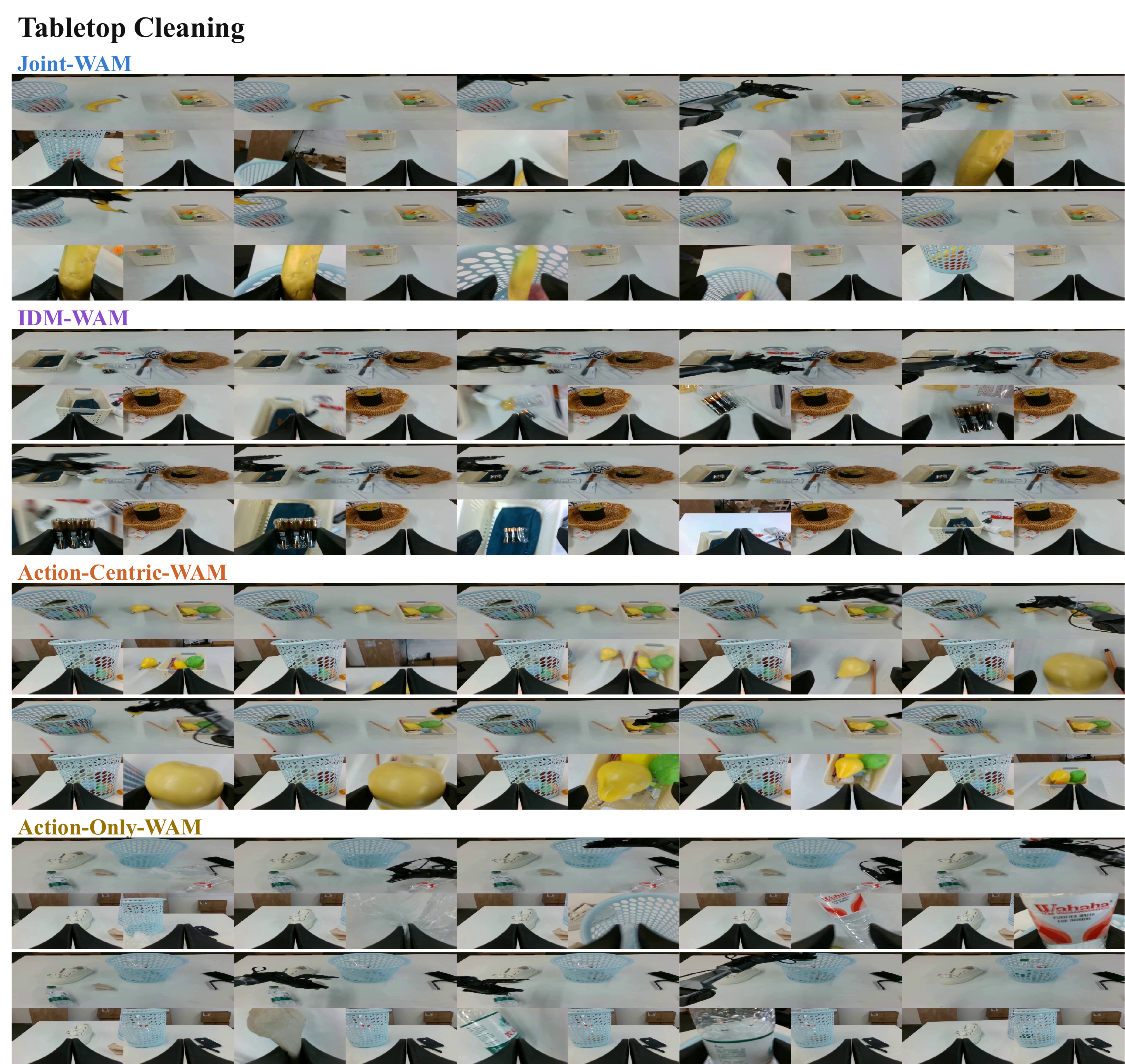}
\caption{Real-world visualizations of Tabletop Cleaning for the four WAM interfaces. The sequences show object pickup, transfer, and placement into a collection basket, with complementary views of reaching and grasping.}
\label{fig:real-world-cleaning}
\end{figure}

Figures~\ref{fig:real-world-clothes}--\ref{fig:real-world-cleaning} present recorded sequences for the three real-world tasks evaluated in Section~\ref{sec:exp-main}, organized by Joint-WAM, IDM-WAM, Action-Centric-WAM, and Action-Only-WAM. Each interface group contains ten selected frames, read from left to right across the first row and then the second. Each frame combines the head-camera view above the two wrist-camera views, showing both the overall scene and local object interactions.

\subsection{Case Study: Routing at Critical Manipulation Steps}
\label{sec:appendix-pair-case-study}

\begin{figure}[H]
\centering
\includegraphics[width=\linewidth]{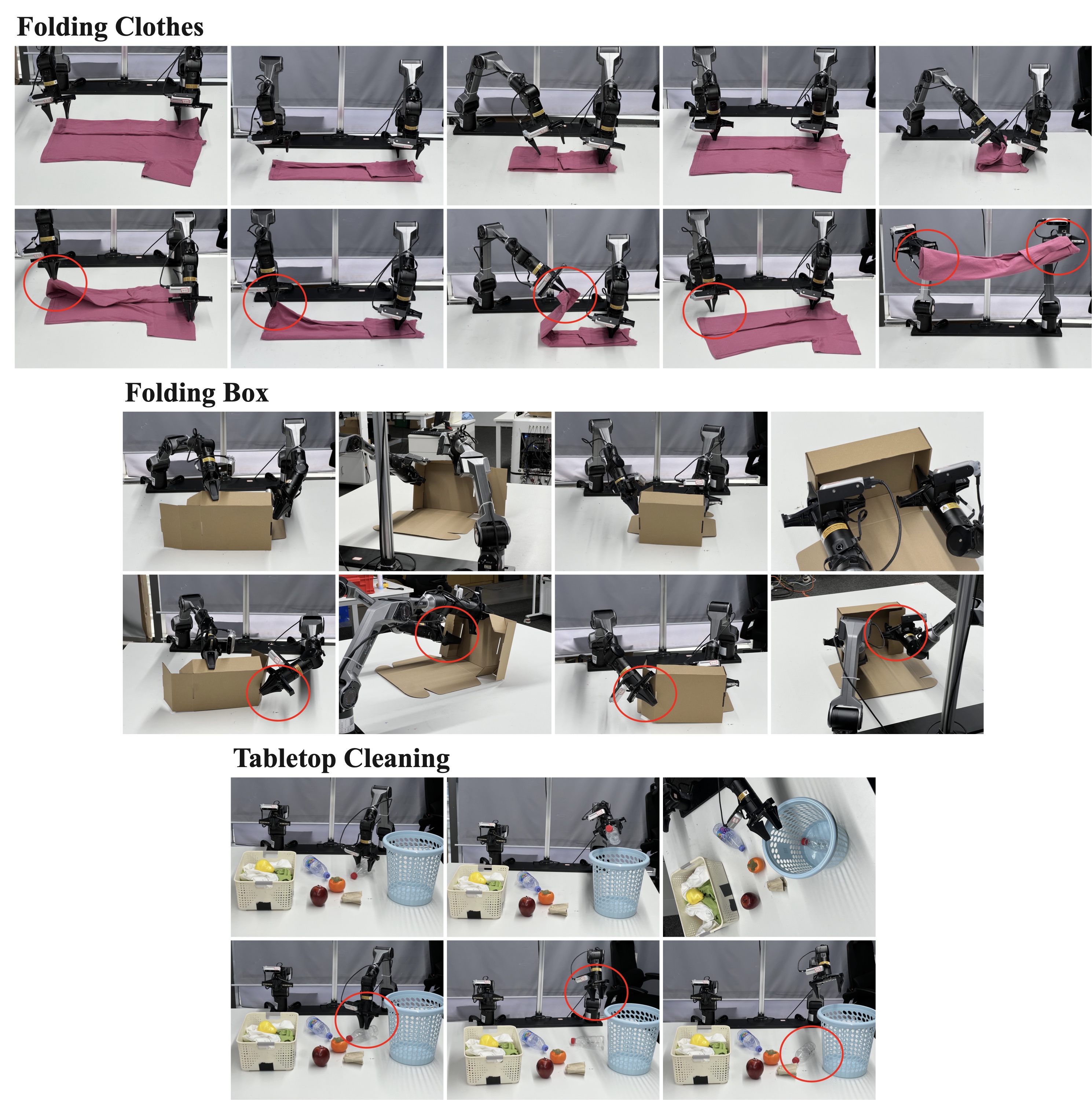}
\caption{Case study of critical operations in Folding Clothes, Folding Box, and Tabletop Cleaning. Within each task group, the upper row shows successful operations and the lower row shows corresponding failure cases, with red circles highlighting the affected interactions. Columns represent distinct cases, read from left to right. The examples illustrate local execution errors that motivate adapting the WAM interface at chunk boundaries.}
\label{fig:pair-case-study}
\end{figure}

Figure~\ref{fig:pair-case-study} illustrates why combining interfaces can benefit real-world task execution. A fixed interface may handle most of a trajectory well yet fail at a critical action. PAIR instead selects an interface for each execution chunk, aiming to exploit the different manipulation capabilities learned by the four interfaces. This allows the choice of action pathway to change with the current interaction, rather than relying on one interface throughout the task.

\paragraph{Folding Clothes.} Small execution errors can disrupt the entire folding sequence. From left to right, the five examples show failures to release the garment during the first, second, and final folding stages; a failed pull-back motion because the gripper does not contact the garment; and lifting the garment again to restart folding after a wrinkled intermediate result. The last behavior constitutes a renewed attempt rather than direct progress toward completion.

\paragraph{Folding Box.} The examples highlight failures to push the left or right side flap into the intended position and complete the corresponding fold. Accurate contact placement and coordinated pushing are essential at these steps.

\paragraph{Tabletop Cleaning.} The three cases show an inaccurate grasp, an object slipping during transfer because the grasp is insecure, and a failure to place the object inside the collection basket.

Across these observed critical operations, Joint-WAM and IDM-WAM handle fine manipulation more reliably and are usually selected by PAIR. These cases suggest how chunk-level selection can reduce the impact of a single interface's local weaknesses on overall task success, consistent with the phase-conditioned preferences in Appendix~\ref{sec:appendix-pair-selection}. The learned router expresses an interface preference from the current context; it does not guarantee an optimal action for every chunk.

\section{Simulation Benchmark Results}
\label{sec:appendix-simulation-results}

This section supplements the real-world comparisons in Section~\ref{sec:exp-main} with simulation benchmark results. The benchmark definitions and evaluation settings are described in Appendix~\ref{sec:appendix-benchmarks}.

\paragraph{LIBERO.}
Table~\ref{tab:libero} reports success rates across the four LIBERO suites. UniW(A)M achieves an average success rate of 97.5\%, with success rates above 95\% in every suite. It obtains the highest Spatial success rate among the compared methods at 99.2\% and exceeds the strongest listed VLA baseline, $\pi_{0.5}$, by 0.6 percentage points in average success. Compared with strong WAM baselines, including Motus, LingBot-VA, and Fast-WAM, UniW(A)M maintains competitive performance while supporting action-conditioned world modeling and multiple action-generation interfaces within a unified framework. These results demonstrate effective action generation in simulation alongside the model's broader predictive capabilities.

\begin{table}[H]
\centering
\footnotesize
\setlength{\tabcolsep}{5pt}
\renewcommand{\arraystretch}{1.05}
\caption{Success rates (\%) on the four LIBERO suites and their arithmetic mean. The best and second-best results in each column are shown in \textbf{bold} and \underline{underlined}, respectively.}
\label{tab:libero}
\begin{tabular}{lccccc}
\toprule
\textbf{Model} & \textbf{Spatial} & \textbf{Object} & \textbf{Goal} & \textbf{Long} & \textbf{Average} \\
\midrule
\rowcolor{gray!15}
\multicolumn{6}{c}{\textit{Vision-Language-Action Models}} \\
OpenVLA~\citep{240609246} & 84.7 & 88.4 & 79.2 & 53.7 & 76.5 \\
$\pi_0$~\citep{241024164} & 96.8 & 98.8 & 95.8 & 85.2 & 94.1 \\
$\pi_{0.5}$~\citep{250416054} & \underline{98.8} & 98.2 & \textbf{98.0} & 92.4 & 96.9 \\
\midrule
\rowcolor{gray!15}
\multicolumn{6}{c}{\textit{World Action Models}} \\
Motus~\citep{251213030} & 96.8 & \underline{99.8} & 96.6 & \underline{97.6} & \underline{97.7} \\
LingBot-VA~\citep{260121998} & 98.5 & 99.6 & \underline{97.2} & \textbf{98.5} & \textbf{98.5} \\
Fast-WAM~\citep{260316666} & 98.2 & \textbf{100.0} & 97.0 & 95.2 & 97.6 \\
\midrule
UniW(A)M & \textbf{99.2} & 99.0 & 96.6 & 95.2 & 97.5 \\
\bottomrule
\end{tabular}
\end{table}

\paragraph{RoboTwin 2.0.}
Table~\ref{tab:robotwin} gives the values visualized in Figure~\ref{fig:robotwin}. UniW(A)M attains 84.38\% C2C success, close to WorldScape Policy 2.0's 84.92\%, and the highest C2R success among the compared methods at 66.60\%. Its C2R score exceeds $\pi_{0.5}$ by 20.60 percentage points, supporting generalization when scene randomization is introduced only at evaluation.

\begin{table}[H]
\centering
\footnotesize %
\setlength{\tabcolsep}{5.0pt} %
\renewcommand{\arraystretch}{0.98} %
\caption{RoboTwin 2.0 success rates (\%) with clean-only training, averaged across 50 tasks separately for C2C and C2R. Official leaderboard baselines~\citep{robotwin-leaderboard-2026} use co-training except $\pi_0$, which uses single-task fine-tuning. Best and second-best scores are \textbf{bold} and \underline{underlined}.}
\label{tab:robotwin}
\begin{tabular}{lcc}
\toprule
\textbf{Model} & \textbf{C2C} & \textbf{C2R} \\
\midrule
$\pi_0$~\citep{241024164} (single-task) & 46.42 & 16.34 \\
$\pi_{0.5}$~\citep{250416054} & 70.7 & \underline{46.0} \\
Fast-WAM~\citep{260316666} & 77.8 & 1.9 \\
WorldScape Policy 2.0~\citep{260718840} & \textbf{84.92} & 38.5 \\
\midrule
UniW(A)M & \underline{84.38} & \textbf{66.60} \\
\bottomrule
\end{tabular}
\end{table}

\section{WorldArena Evaluation and Future-World Prediction}
\label{sec:appendix-worldarena}

\subsection{Results by Evaluation Metric}
\label{sec:appendix-worldarena-metrics}

Tables~\ref{tab:worldarena-visual-motion-content} and~\ref{tab:worldarena-physics-3d-control} report the complete set of component scores for the three real-world tasks evaluated with WorldArena, grouped into six evaluation dimensions. We retain the original $0$--$1$ scale and follow the task and model order in Table~\ref{tab:worldarena-quality}, which reports the aggregate EWMScore-P scores.

\begin{table}[H]
\centering
\caption{WorldArena component scores for visual quality, motion quality, and content consistency. The highest score for each metric within each task is shown in \textbf{bold}, including ties.}
\label{tab:worldarena-visual-motion-content}
\scriptsize
\setlength{\tabcolsep}{3pt}
\renewcommand{\arraystretch}{1.15}
\begin{tabular*}{\textwidth}{@{\extracolsep{\fill}}lccccccccc}
\toprule
\textbf{Model} & \multicolumn{3}{c}{\textbf{Visual Quality}} & \multicolumn{3}{c}{\textbf{Motion Quality}} & \multicolumn{3}{c}{\textbf{Content Consistency}} \\
\cmidrule(lr){2-4}\cmidrule(lr){5-7}\cmidrule(lr){8-10}
 & \shortstack{Image\\Quality} & \shortstack{Aesthetic\\Quality} & \shortstack{JEPA\\Similarity} & \shortstack{Dynamic\\Degree} & \shortstack{Flow\\Score} & \shortstack{Motion\\Smoothness} & \shortstack{Subject\\Consistency} & \shortstack{Backgrd.\\Consistency} & \shortstack{Photom.\\Consistency} \\
\midrule
\rowcolor{gray!15}
\multicolumn{10}{c}{\textit{Task 1: Folding Clothes (Deformable)}} \\
\shortstack[l]{Cosmos-Predict 2.5\\\citep{251100062}} & \textbf{0.3623} & 0.2389 & 0.2186 & 0.3057 & 0.1940 & 0.6070 & \textbf{0.7122} & \textbf{0.8976} & \textbf{0.2339} \\
CtrlWorld~\citep{251010125} & 0.3237 & \textbf{0.2549} & 0.1757 & 0.6430 & 0.6827 & 0.7821 & 0.6917 & 0.8932 & 0.0469 \\
\textbf{UniW(A)M (Ours)} & 0.3158 & 0.2189 & \textbf{0.4439} & \textbf{0.6778} & \textbf{0.8025} & \textbf{0.8777} & 0.6924 & 0.8772 & 0.0177 \\
\midrule
\rowcolor{gray!15}
\multicolumn{10}{c}{\textit{Task 2: Folding Box (Contact-Rich)}} \\
\shortstack[l]{Cosmos-Predict 2.5\\\citep{251100062}} & \textbf{0.3414} & 0.2403 & 0.4027 & 0.1439 & 0.0847 & 0.5247 & 0.6797 & 0.7887 & \textbf{0.3798} \\
CtrlWorld~\citep{251010125} & 0.3179 & \textbf{0.2442} & 0.6381 & \textbf{0.4236} & \textbf{0.2647} & 0.6656 & \textbf{0.7407} & 0.8795 & 0.1597 \\
\textbf{UniW(A)M (Ours)} & 0.3235 & 0.2373 & \textbf{0.6541} & 0.3755 & 0.2290 & \textbf{0.6869} & 0.7396 & \textbf{0.8845} & 0.1459 \\
\midrule
\rowcolor{gray!15}
\multicolumn{10}{c}{\textit{Task 3: Tabletop Clean. (Long-Horizon)}} \\
\shortstack[l]{Cosmos-Predict 2.5\\\citep{251100062}} & \textbf{0.4224} & 0.2264 & 0.0463 & 0.0922 & 0.0318 & 0.4849 & 0.1672 & 0.1885 & 0.2691 \\
CtrlWorld~\citep{251010125} & 0.3472 & \textbf{0.3696} & 0.3376 & \textbf{0.3118} & \textbf{0.1329} & 0.5793 & 0.8706 & 0.9350 & \textbf{0.3125} \\
\textbf{UniW(A)M (Ours)} & 0.3838 & 0.3681 & \textbf{0.5030} & 0.2742 & 0.1083 & \textbf{0.6310} & \textbf{0.8914} & \textbf{0.9353} & 0.2769 \\
\bottomrule
\end{tabular*}
\end{table}

\begin{table}[H]
\centering
\caption{WorldArena component scores for physics adherence, 3D accuracy, and controllability. The highest score for each metric within each task is shown in \textbf{bold}, including ties.}
\label{tab:worldarena-physics-3d-control}
\scriptsize
\setlength{\tabcolsep}{3pt}
\renewcommand{\arraystretch}{1.15}
\begin{tabular*}{\textwidth}{@{\extracolsep{\fill}}lcccccc}
\toprule
\textbf{Model} & \multicolumn{2}{c}{\textbf{Physics Adherence}} & \multicolumn{2}{c}{\textbf{3D Accuracy}} & \multicolumn{2}{c}{\textbf{Controllability}} \\
\cmidrule(lr){2-3}\cmidrule(lr){4-5}\cmidrule(lr){6-7}
 & \shortstack{Interaction\\Quality} & \shortstack{Trajectory\\Accuracy} & \shortstack{Depth\\Accuracy} & \shortstack{Perspectivity} & \shortstack{Instruction\\Following} & \shortstack{Semantic\\Alignment} \\
\midrule
\rowcolor{gray!15}
\multicolumn{7}{c}{\textit{Task 1: Folding Clothes (Deformable)}} \\
\shortstack[l]{Cosmos-Predict 2.5\\\citep{251100062}} & 0.6400 & 0.0720 & 0.5474 & \textbf{0.8000} & 0.5650 & 0.8175 \\
CtrlWorld~\citep{251010125} & 0.6350 & 0.1856 & 0.4754 & \textbf{0.8000} & 0.5150 & 0.8593 \\
\textbf{UniW(A)M (Ours)} & \textbf{0.8000} & \textbf{0.2287} & \textbf{0.6460} & \textbf{0.8000} & \textbf{0.9350} & \textbf{0.8929} \\
\midrule
\rowcolor{gray!15}
\multicolumn{7}{c}{\textit{Task 2: Folding Box (Contact-Rich)}} \\
\shortstack[l]{Cosmos-Predict 2.5\\\citep{251100062}} & \textbf{0.7650} & 0.1091 & 0.6825 & \textbf{0.8150} & \textbf{0.9500} & \textbf{0.7537} \\
CtrlWorld~\citep{251010125} & 0.6450 & 0.1473 & \textbf{0.8754} & 0.8000 & 0.7800 & 0.7407 \\
\textbf{UniW(A)M (Ours)} & 0.7050 & \textbf{0.1728} & 0.8464 & 0.8000 & 0.8400 & 0.7431 \\
\midrule
\rowcolor{gray!15}
\multicolumn{7}{c}{\textit{Task 3: Tabletop Clean. (Long-Horizon)}} \\
\shortstack[l]{Cosmos-Predict 2.5\\\citep{251100062}} & 0.5400 & 0.0288 & 0.5741 & 0.5800 & 0.4100 & 0.7009 \\
CtrlWorld~\citep{251010125} & 0.6000 & 0.0594 & 0.5653 & \textbf{0.8000} & 0.5900 & \textbf{0.7234} \\
\textbf{UniW(A)M (Ours)} & \textbf{0.6650} & \textbf{0.0726} & \textbf{0.5773} & \textbf{0.8000} & \textbf{0.6850} & 0.7077 \\
\bottomrule
\end{tabular*}
\end{table}

\subsection{Qualitative Comparison of Future-World Prediction}
\label{sec:appendix-world-prediction-qualitative}

\begin{figure}[H]
\centering
\includegraphics[width=\linewidth]{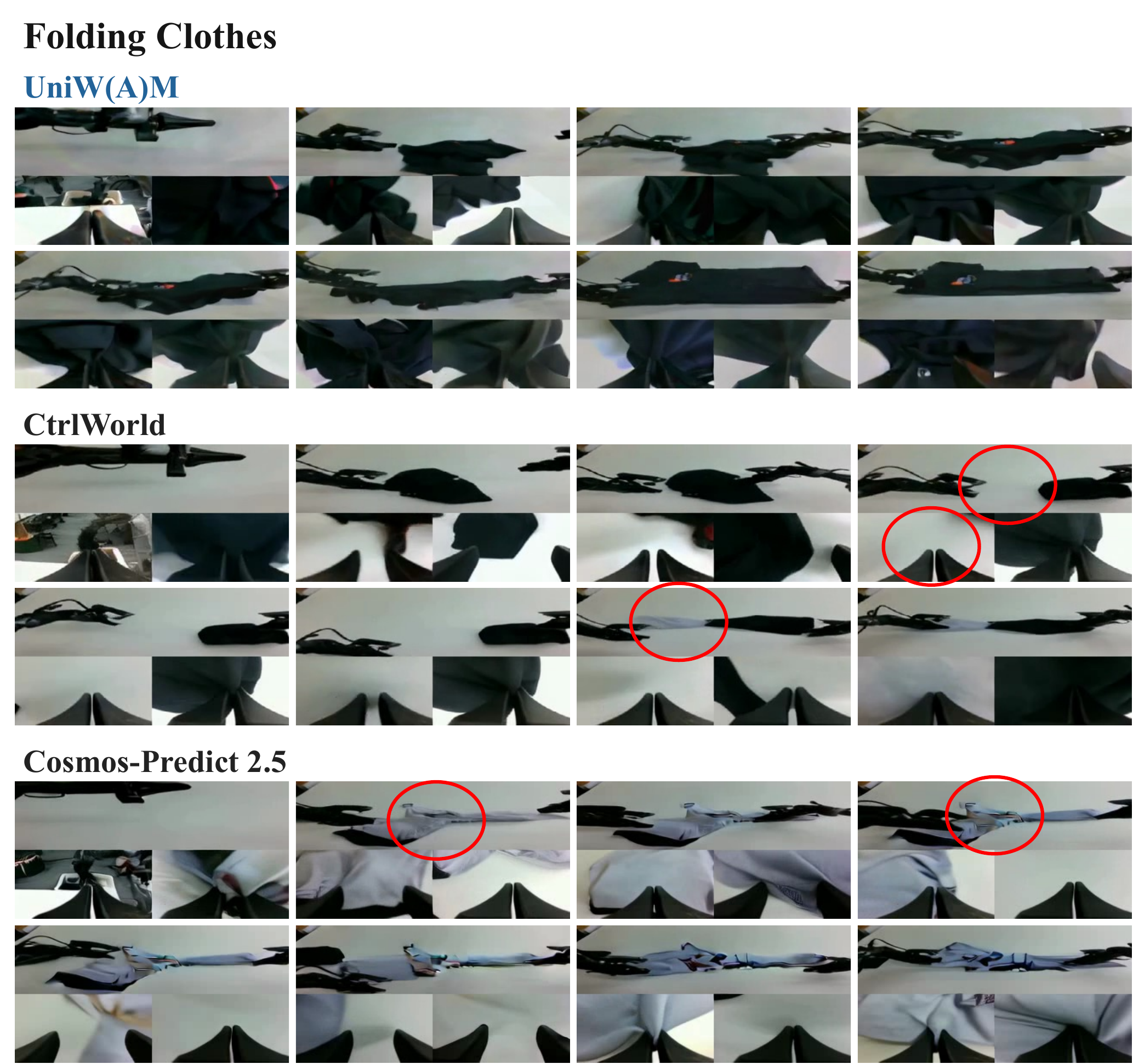}
\caption{Action-conditioned future-world prediction on Folding Clothes. UniW(A)M in AC-WM mode better preserves garment visibility and appearance in this early-sequence comparison. Red circles highlight missing garment regions and inconsistent fabric appearance in the baselines.}
\label{fig:acwm-clothes}
\end{figure}

\begin{figure}[H]
\centering
\includegraphics[width=\linewidth]{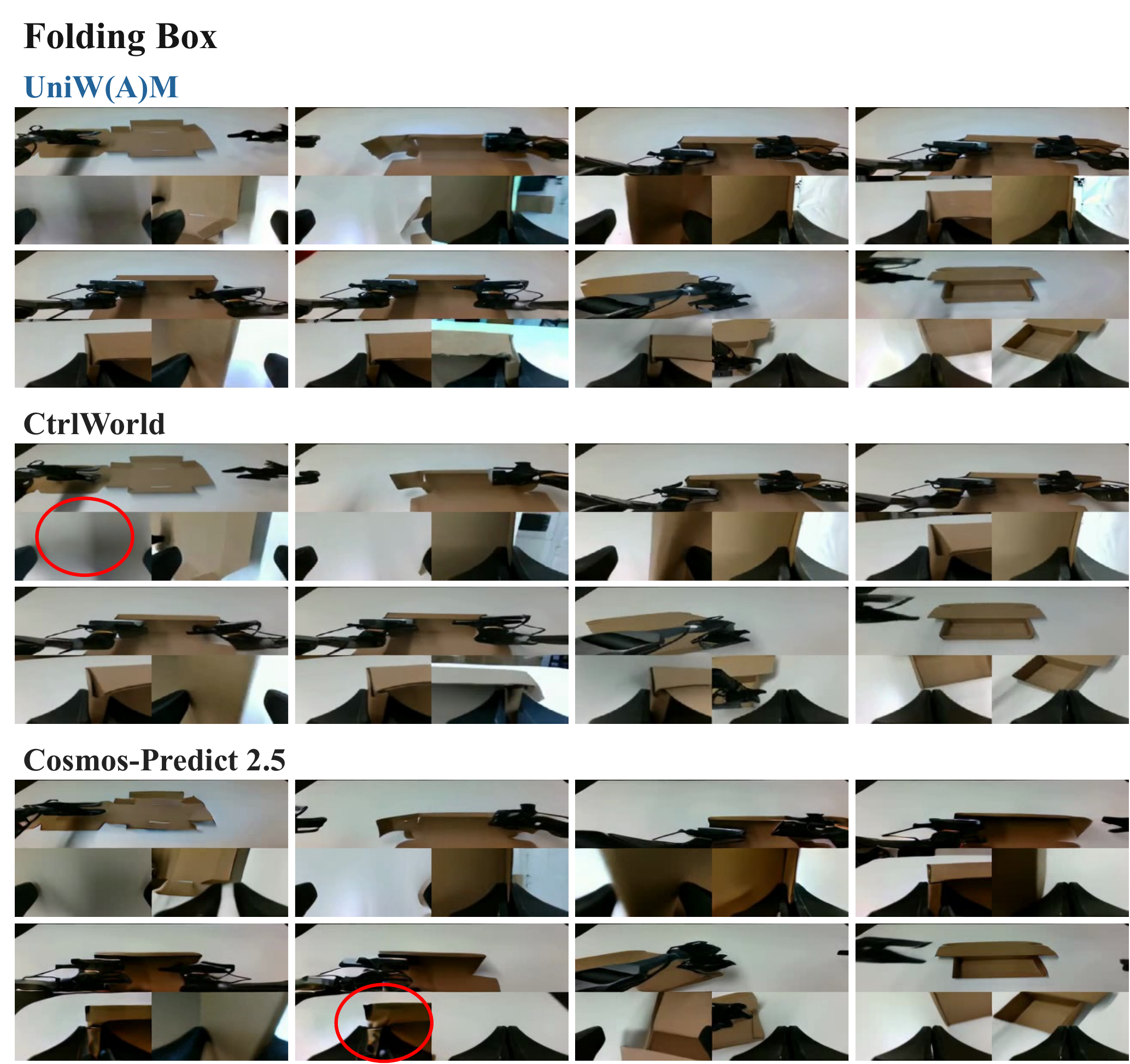}
\caption{Action-conditioned future-world prediction on Folding Box. The comparison emphasizes cardboard structure and wrist-view contact details. Red circles identify localized blur and poorly resolved boundaries in baseline predictions.}
\label{fig:acwm-box}
\end{figure}

\begin{figure}[H]
\centering
\includegraphics[width=\linewidth]{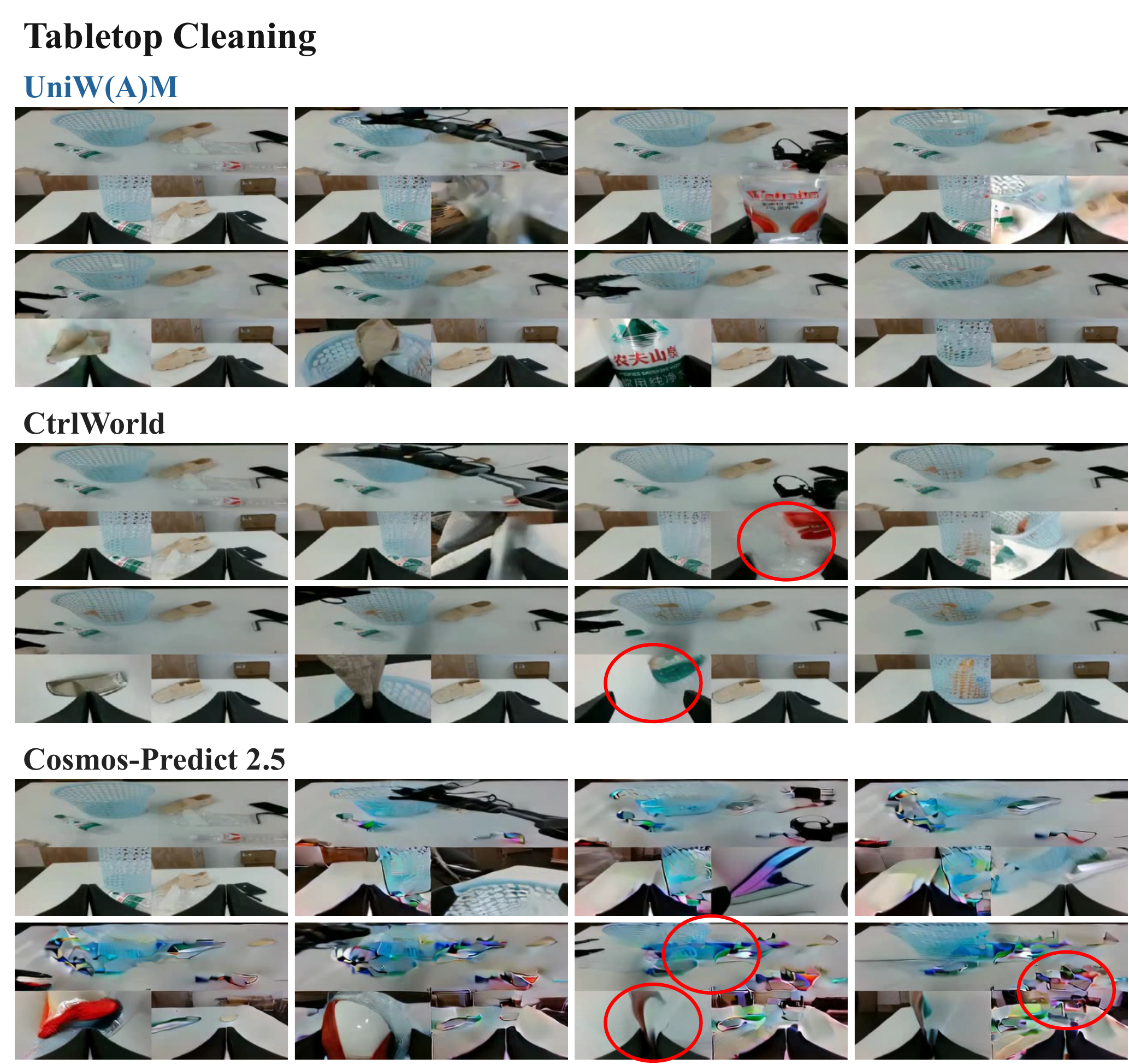}
\caption{Action-conditioned future-world prediction on Tabletop Cleaning. UniW(A)M retains more recognizable bottle-label details and scene appearance. Red circles highlight lost label texture, color drift, and distorted object or background regions in the baselines.}
\label{fig:acwm-cleaning}
\end{figure}

We compare videos generated by UniW(A)M in its AC-WM mode with CtrlWorld~\citep{251010125} and Cosmos-Predict 2.5~\citep{251100062} to examine whether unified training preserves action-conditioned scene evolution and fine visual details. Figures~\ref{fig:acwm-clothes}--\ref{fig:acwm-cleaning} present three task-specific comparisons. Each model has eight frames arranged in two rows of four, read left to right and then top to bottom; corresponding positions use matched video timestamps. Red circles mark selected prediction errors in the baselines. These qualitative examples complement the component scores by exposing differences in object persistence, appearance consistency, and local interaction details.

\paragraph{Folding Clothes.} In the highlighted CtrlWorld predictions, portions of the garment disappear from the head or wrist view, and different regions of the garment acquire inconsistent colors. Cosmos-Predict 2.5 changes the dark garment to a pale, bluish appearance and distorts its local folds. UniW(A)M more consistently retains the dark fabric, a connected garment silhouette, and visible folds around the grippers. Preserving these features makes the predicted deformation easier to follow as the arms manipulate the cloth.

\paragraph{Folding Box.} The three methods capture a broadly similar progression from flat cardboard to a folded box, while differing in local image quality. The highlighted CtrlWorld wrist view contains substantial blurring near the cardboard, obscuring its structure. In the marked Cosmos-Predict 2.5 prediction, darkened and distorted local regions make the boundary between the box flap and the gripper less distinct. UniW(A)M retains more discernible cardboard edges and gripper--object boundaries at the corresponding moments, providing a clearer depiction of the contact geometry during folding. The contrast here is primarily in local detail, rather than a different overall task sequence.

\paragraph{Tabletop Cleaning.} Bottle pickup and transfer expose differences in fine texture and object identity. UniW(A)M preserves more recognizable red lettering and label structure on the bottles as they approach the wrist cameras. CtrlWorld retains the coarse bottle shape but loses or smears much of the label detail in the highlighted frames. Cosmos-Predict 2.5 exhibits stronger color drift and shape distortion, affecting both manipulated objects and the surrounding scene; some regions develop vivid, fragmented textures. By maintaining more stable bottle appearance, basket structure, and background content, UniW(A)M provides a more coherent visual account of the manipulation. Together, these selected comparisons support the unified framework's ability to retain useful future-world modeling capabilities alongside action generation.

\section{Training Details and Ablation Studies}
\label{sec:appendix-ablation}

\subsection{Uniform Interface Sampling}
\label{sec:appendix-sampling}

During joint training, the five interfaces are sampled uniformly with probability $0.2$ each, corresponding to the ratio $1{:}1{:}1{:}1{:}1$, without a manually tuned sampling schedule. For the four-WAM-only ablation, AC-WM is omitted and the remaining interfaces are sampled uniformly with probability $0.25$ each. With the total number of optimization steps fixed, each WAM interface therefore receives 25\% more updates in expectation than under five-paradigm sampling. These sampling probabilities are distinct from the relative weighting of video and action losses within each interface, specified below.

\subsection{Interface-Specific Training Objectives}
\label{sec:appendix-training-objectives}

\paragraph{Shared flow-matching formulation.}
All five interfaces use flow matching, with their conditioning pathways determined by the visibility masks in Appendix~\ref{sec:appendix-interface-masks}. Let $x_{\mathrm{vid}}$ denote future video latents produced by the video encoder and $x_{\mathrm{act}}$ the corresponding action sequence. For each modality $m\in\{\mathrm{vid},\mathrm{act}\}$ to be predicted, Gaussian noise is interpolated with the training target:
\begin{equation}
    x_{m,\sigma_m}=(1-\sigma_m)\odot x_m+\sigma_m\odot\epsilon_m,
    \qquad \epsilon_m\sim\mathcal{N}(0,I),
    \qquad u_m^{\star}=\epsilon_m-x_m.
    \label{eq:appendix-flow-interpolation}
\end{equation}
Here, $\odot$ denotes elementwise multiplication with broadcasting over feature dimensions. Noise levels are sampled per temporal chunk and shared by the video and action targets within that chunk. The training scheduler samples uniformly over 1,000 discrete indices and applies a shifted noise schedule with shift parameter $5$; thus, the resulting noise levels $\sigma_m$ are not uniformly distributed on $[0,1]$.

For interface $p$, let $\hat u_{\theta,m,j}^{(p)}$ be its predicted velocity for target unit $j$, conditioned on the available trajectory context and the interface-specific mask. A target unit is a video latent frame or an action vector. We define the modality loss as
\begin{equation}
    \mathcal{L}_{m}^{(p)}
    =\mathbb{E}\!\left[
    \frac{1}{N_m D_m}\sum_{j=1}^{N_m}
    w(\sigma_{m,j})
    \left\|b_{m,j}\odot
    \bigl(\hat u_{\theta,m,j}^{(p)}-u_{m,j}^{\star}\bigr)\right\|_2^2
    \right].
    \label{eq:appendix-modality-flow-loss}
\end{equation}
The expectation covers training trajectories, Gaussian noise, and sampled noise levels. Here, $N_m$ is the number of target units, $D_m$ their feature dimensionality, and $w$ the scheduler's time-dependent loss weight. The binary mask $b_{m,j}$ excludes unavailable actions and invalid action dimensions; it is identically one for video targets. The clean initial video frame is excluded from the video target set. Both modality losses average over their respective target elements before being combined.

\paragraph{Joint-WAM and IDM-WAM.}
These interfaces use the same loss composition:
\begin{equation}
    \mathcal{L}^{(p)}
    =\mathcal{L}_{\mathrm{vid}}^{(p)}+\mathcal{L}_{\mathrm{act}}^{(p)},
    \qquad p\in\{\text{Joint-WAM},\text{IDM-WAM}\}.
    \label{eq:appendix-joint-idm-objective}
\end{equation}
Their conditional prediction problems nevertheless differ. Joint-WAM permits joint video--action denoising, whereas IDM-WAM uses interleaved teacher forcing: action denoising can access the preceding video condition under the video-to-action ordering. IDM-WAM maintains separate target and conditioning copies, and only the noisy target copies receive denoising supervision. With probability $0.5$ per training forward pass, its video conditioning copies receive additional noise augmentation; action conditioning copies remain clean.

\paragraph{Action-Centric-WAM and Action-Only-WAM.}
Both interfaces retain video supervision while placing greater weight on action prediction:
\begin{equation}
    \begin{aligned}
        \mathcal{L}^{(p)}
        &=\mathcal{L}_{\mathrm{vid}}^{(p)}
        +\lambda_{\mathrm{act}}\mathcal{L}_{\mathrm{act}}^{(p)},
        \qquad \lambda_{\mathrm{act}}=5,\\
        p&\in\{\text{Action-Centric-WAM},\text{Action-Only-WAM}\}.
    \end{aligned}
    \label{eq:appendix-action-objective}
\end{equation}
In Action-Centric-WAM, the video branch attends to the current noisy action sequence, while action prediction does not access the current future-video targets. Action-Only-WAM also blocks the action-to-video direction, so video prediction provides auxiliary supervision without current-chunk cross-modal attention. Both objectives include video loss during training. At deployment, neither interface predicts future video. WM is inactive in Action-Centric-WAM, whereas VGM remains active for the initial conditioning frame in Action-Only-WAM; shared and VLA experts process future action tokens in both interfaces.

\paragraph{AC-WM.}
This interface predicts future video with the ground-truth action sequence supplied as a clean condition:
\begin{equation}
    \mathcal{L}^{(\mathrm{AC\text{-}WM})}
    =\mathcal{L}_{\mathrm{vid}}^{(\mathrm{AC\text{-}WM})}.
    \label{eq:appendix-acwm-objective}
\end{equation}
Only the future video targets are noised and supervised. The action sequence is neither corrupted nor assigned an action-denoising loss.

Combining these objectives with uniform interface sampling yields
\begin{equation}
    \mathcal{L}_{\mathrm{train}}
    =\frac{1}{5}\sum_{p\in\mathcal{P}}\mathcal{L}^{(p)},
    \label{eq:appendix-mixed-training-objective}
\end{equation}
where $\mathcal{P}$ contains the five interfaces above. Each optimization step uses the loss of one sampled interface, providing a stochastic estimate of this mixed objective.

\subsection{Initialization and Optimization Settings}
\label{sec:appendix-expert-initialization}

Each MoE sublayer is initialized from its corresponding pretrained dense feed-forward network. Its parameters are copied into five independently trainable experts: shared, VGM, VLA, WM, and IDM. The bias-free router matrix $W_r$ is initialized to zero, assigning equal initial weights to the experts allowed for each group. Because these weights sum to one and every expert initially implements the same dense function, the replacement preserves the feed-forward output at initialization:
\begin{equation}
    \operatorname{MoE}^{(0)}(h)
    =\sum_{e\in\mathcal{E}_g}\frac{1}{|\mathcal{E}_g|}f_{\mathrm{dense}}(h)
    =f_{\mathrm{dense}}(h).
    \label{eq:appendix-expert-initialization}
\end{equation}
This identity also holds for shared-only groups and is understood up to numerical precision. It concerns the feed-forward replacement under identical inputs, not equivalence between interfaces with different attention masks. Subsequent training updates the copied experts independently, while the learned router adjusts their relative contributions within the permitted sets.

\paragraph{Optimization settings.}
Table~\ref{tab:training-configuration} summarizes the optimization settings for backbone training and PAIR adaptation. The backbone batch size is the effective global batch across devices and gradient accumulation. Each PAIR classification head is trained independently with a batch size of 128; this batch size is not multiplied by the number of ensemble members. The backbone weight decay applies to the decay parameter group, while parameters assigned to the no-decay group, including biases, use zero weight decay.

\begin{table}[H]
\centering
\caption{Optimization settings for backbone training and PAIR classification-head training.}
\label{tab:training-configuration}
\begin{tabular}{lcc}
\toprule
Configuration & Backbone & PAIR classification head \\
\midrule
Optimizer & AdamW & AdamW \\
Learning rate & $1\times10^{-5}$ & $1\times10^{-3}$ \\
Learning-rate schedule & Constant & Constant, no warm-up \\
Weight decay & $1\times10^{-5}$ & $1\times10^{-2}$ \\
Global batch size & 192 & 128 per head \\
\bottomrule
\end{tabular}
\end{table}

\subsection{Architecture, Routing, and Training-Paradigm Ablations}
\label{sec:appendix-architecture-ablation}

Table~\ref{tab:ablate-mix} compares a single dense DiT, MoE with sample-level routing, and UniW(A)M with CATR. All three use the same training steps, learning-rate settings, and training data volume. The single DiT is initialized from Wan2.1-I2V-14B-480P~\citep{250320314}. We report its approximate scale as 16.4B parameters based on the pretrained DiT, omitting the action/state adapters from this estimate. The MoE configurations use a Wan2.2-TI2V-5B backbone with five FFN experts per block and approximately 16.2B parameters. Both figures exclude external encoders and VAE. This is not a strictly parameter-matched comparison, and backbone configurations and initialization sources also differ. In contrast, sample-level routing and CATR use identical expert counts, network structures, and parameter counts. They differ in routing organization: CATR assigns eligible expert sets and learned mixture weights to modality- and role-specific token groups, whereas sample-level routing applies the active expert ensemble to the entire sample. This comparison evaluates the combined token-group routing design, rather than isolating pooling granularity alone.

\begin{table}[H]
\centering
\caption{Architecture, routing, and training-paradigm ablations on three real-world tasks. The dense, sample-level routing, and full five-paradigm CATR configurations use the same training steps, learning-rate settings, and training data volume; the latter two have identical expert architectures and parameter counts. The four-WAM-only CATR variant shares the full model's initialization, training steps, learning-rate settings, batch size, and dataset, but samples only the four WAMs uniformly. Success rates are reported in percentage (\%).}
\label{tab:ablate-mix}
\footnotesize
\setlength{\tabcolsep}{5.0pt}
\renewcommand{\arraystretch}{1.08}
\begin{tabular}{lcccc}
\toprule
\multirow{2}{*}{\textbf{Interface}} & \textbf{Folding Clothes} & \textbf{Folding Box} & \textbf{Tabletop Clean.} & \multirow{2}{*}{\textbf{Average}} \\
& \scriptsize\textit{(Deformable)} & \scriptsize\textit{(Contact-Rich)} & \scriptsize\textit{(Long-Horizon)} & \\
\midrule
\rowcolor{gray!15}
\multicolumn{5}{c}{\textit{Single DiT (Dense)}} \\
Joint-WAM           & 10 & 10 & 20 & 13.3 \\
IDM-WAM             & 20 & 0 & 30 & 16.7 \\
Action-Centric-WAM  & 0  & 0 & 10 & 3.3 \\
Action-Only-WAM     & 0  & 0 & 5 & 1.7 \\
\rowcolor{gray!15}
\multicolumn{5}{c}{\textit{MoE with Sample-Level Routing}} \\
Joint-WAM           & 50 & 45 & 60 & 51.7 \\
IDM-WAM             & 80 & 30 & 70 & 60.0 \\
Action-Centric-WAM  & 10 & 0 & 20 & 10.0 \\
Action-Only-WAM     & 10 & 0 & 20 & 10.0 \\
\rowcolor{gray!15}
\multicolumn{5}{c}{\textit{UniW(A)M with CATR: Four WAMs Only (w/o AC-WM)}} \\
Joint-WAM           & 75 & 60 & 80 & 71.7 \\
IDM-WAM             & 80 & 60 & 80 & 73.3 \\
Action-Centric-WAM  & 55 & 30 & 65 & 50.0 \\
Action-Only-WAM     & 50 & 10 & 50 & 36.7 \\
\rowcolor{gray!15}
\multicolumn{5}{c}{\textit{UniW(A)M with CATR: All Five Paradigms (Ours)}} \\
Joint-WAM           & \textbf{90} & 80 & \textbf{90} & 86.7 \\
IDM-WAM             & \textbf{90} & \textbf{85} & \textbf{90} & \textbf{88.3} \\
Action-Centric-WAM  & 85 & 75 & 85 & 81.7 \\
Action-Only-WAM     & 85 & 70 & 85 & 80.0 \\
\bottomrule
\end{tabular}
\end{table}

The dense configuration reaches at most 16.7\% average success. MoE with sample-level routing improves Joint-WAM and IDM-WAM to 51.7\% and 60.0\%, but both action-only inference pathways remain at 10.0\%. CATR raises the four interfaces to 80.0\%--88.3\%, including 81.7\% and 80.0\% for Action-Centric-WAM and Action-Only-WAM. Thus, an expert architecture alone is insufficient to sustain all interfaces in this setting, while token-group alignment improves joint-training performance. These comparisons do not isolate the contributions of gradient interference or other optimization mechanisms.

\paragraph{Contribution of AC-WM co-training.} The four-WAM-only variant retains UniW(A)M with CATR but omits AC-WM from the training paradigm set. It uses the same initialization, total training steps, learning-rate settings, batch size, and dataset as the full five-paradigm model. Each retained interface is sampled with probability $0.25$, compared with $0.2$ in the full model. Thus, the full model's gains occur despite allocating fewer expected training updates to each WAM interface. As shown in Table~\ref{tab:ablate-mix}, its average success rates are 71.7\%, 73.3\%, 50.0\%, and 36.7\% for Joint-WAM, IDM-WAM, Action-Centric-WAM, and Action-Only-WAM, respectively, compared with 86.7\%, 88.3\%, 81.7\%, and 80.0\% under full five-paradigm training. Every interface--task pair improves when AC-WM is included. The largest task-level gains occur on box folding: Action-Centric-WAM increases from 30\% to 75\%, and Action-Only-WAM from 10\% to 70\%. These interfaces omit future-video prediction at deployment, so the observed benefit is not restricted to execution paths that explicitly generate visual rollouts. The comparison supports a positive contribution of AC-WM co-training to WAM action generation, while not isolating the optimization mechanism or establishing the reverse effect of WAM training on AC-WM quality.

\end{document}